\documentclass[sigconf,nonacm]{acmart}

\AtBeginDocument{%
  \providecommand\BibTeX{{%
    \normalfont B\kern-0.5em{\scshape i\kern-0.25em b}\kern-0.8em\TeX}}}

\setcopyright{none}
\renewcommand\footnotetextcopyrightpermission[1]{}

\usepackage{graphicx}
\usepackage{amsmath}
\usepackage{algorithm}
\usepackage{algpseudocode}
\usepackage{makecell}
\usepackage{array}
\usepackage{tablefootnote}
\usepackage{adjustbox}
\usepackage{hyperref}
\usepackage{breqn}

\usepackage{amssymb}
\usepackage{booktabs}
\usepackage{wrapfig}
\usepackage{caption}
\usepackage{subcaption}
\usepackage{float}
\usepackage{xcolor}

\title{EUPHORIA: Efficient Universal Planning via Hybrid Optimization for Robust Industrial Robotic Assembly}

\author{%
  Shih-Yu Lai\textsuperscript{1,2},~
  Chia-Ching Yen\textsuperscript{3},~
  Yang-Ting Shen\textsuperscript{3},~
  Peter Yichen Chen\textsuperscript{4},~
  Yu-Lun Liu\textsuperscript{5},~
  Bing-Yu Chen\textsuperscript{1$\dagger$}%
}
\affiliation{%
  \institution{%
    \textsuperscript{1}National Taiwan University\quad
    \textsuperscript{2}MoonShine Animation Studio\quad
    \textsuperscript{3}National Cheng Kung University\\
    \textsuperscript{4}The University of British Columbia\quad
    \textsuperscript{5}National Yang Ming Chiao Tung University%
  }%
  \country{~}
  \city{~}
}

\begin{teaserfigure}
  \centering
  \vspace{-1.0em}
  \includegraphics[width=\textwidth]{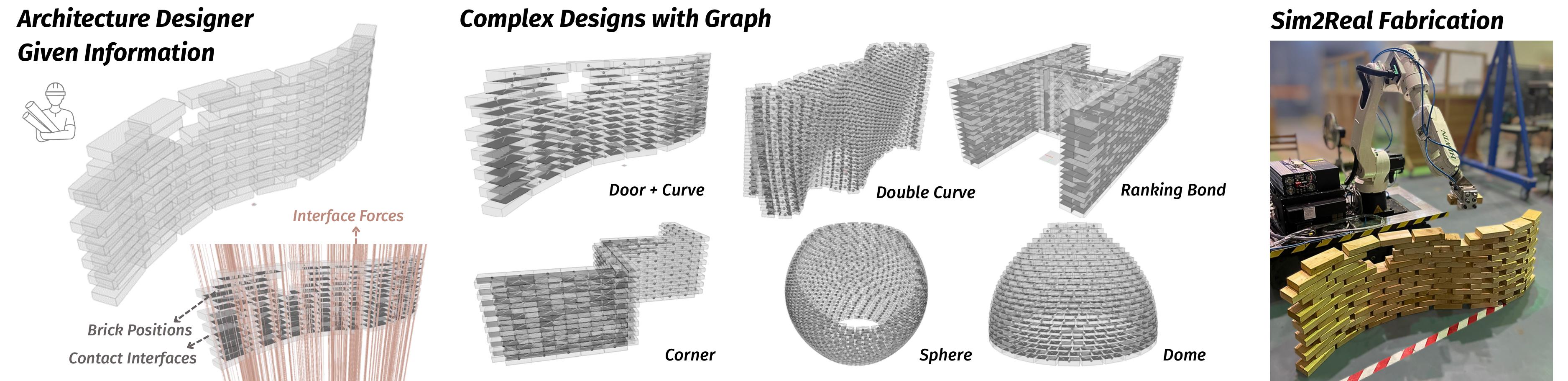}
  \vspace{-1.0em}
  \caption{We present \textsc{Euphoria}, a robotic assembly system that closes the gap between parametric architectural design and physical construction. Unlike prior learning-based planners that require per-geometry retraining and decouple sequencing from motion, a designer-authored brick CAD with contact interfaces and DEM-derived forces (\textit{left}) is realized by a mobile manipulator (\textit{right}) via a \emph{physics-informed and design-aware} planner: \textbf{1.~Universal}---a \emph{single} model few-shot generalization adapts to unseen customized geometries (curved walls, ranking bonds, domes, spheres, etc.) without retraining (\textit{middle}); \textbf{2.~Efficient}---kinematics-aware sequencing yields low-energy, smooth trajectories; \textbf{3.~Robust}---residual correction ensures millimeter-scale Sim2Real accuracy.}
  \label{fig:teaser}
\end{teaserfigure}
\vspace{-1.5em}

\begin{abstract}
Robotic assembly in architectural construction faces a persistent bottleneck: existing planners are either highly specialized—requiring computationally prohibitive retraining for every novel geometric design—or operationally inefficient, treating structural sequencing and kinematic motion generation as disjoint processes. In this work, we present EUPHORIA, a unified framework that achieves universal few-shot adaptability and dynamic efficiency through the hybrid optimization strategy. To overcome the "retraining bottleneck," we propose a \textbf{Meta-Geometric Encoder based on Graph Hypernetworks}. Unlike standard contrastive learning that performs only feature-level recognition, our hypernetwork dynamically generates policy parameters from a minimal support set, enabling few-shot parameter-level adaptation to complex topologies (e.g., domes, arches) without gradient-based retraining. For structural reasoning, we introduce a \textbf{Physics-Informed Graph Transformer trained via Soft Actor-Critic (SAC)}. Unlike standard graph policies, we implement a Physics-Bias Attention Mechanism that directly modulates attention scores using discrete contact forces derived from Discrete Element Model (DEM) simulations, guiding the planner to prioritize structurally critical connections. Furthermore, we ensure operational efficiency through Kinematics-Aware Sequencing, where the SAC objective explicitly penalizes high-energy transitions. Finally, we \textbf{bridge the Sim2Real gap via Residual Stability Correction}, a differentiable optimization layer that fine-tunes coarse assembly actions by minimizing a joint energy-stability cost function prior to execution. Extensive experiments demonstrate that EUPHORIA significantly reduces energy consumption compared to decoupled baselines and achieves state-of-the-art success rates on unseen, non-standard geometries with minimal few-shot examples, effectively bridging the gap between digital design and physical construction. This approach fundamentally redefines robotic assembly by fusing meta-learning, physics-informed attention, and residual optimization into a cohesive, generalized planner.
\end{abstract}

\begin{document}
\maketitle
\renewcommand{\thefootnote}{}
\footnotetext{$^\dagger$Corresponding author.\quad
  \textit{Emails:}~\texttt{akinesia112@gmail.com},~\texttt{ccyen@gs.ncku.edu.tw},\\
  \texttt{bowbowshen@gmail.com},
  ~\texttt{pyc@csail.mit.edu},~%
  \texttt{yulunliu@cs.nycu.edu.tw},~\texttt{robin@ntu.edu.tw}}
\setcounter{footnote}{0}
\renewcommand{\thefootnote}{\arabic{footnote}}

\section{Introduction}

The architectural domain is shifting from standardized mass production to \textbf{mass customization}, increasing demand for industrial robotic systems that can assemble bespoke structures such as parametric curved walls, domes, and arches \cite{Intro0, Intro1, Intro2}. Yet realizing these designs remains challenging: trial-and-error brick assembly with physical robots is costly in time, materials, and hardware wear, making a "Simulator-First" workflow—validating stability and feasibility in a high-fidelity digital twin before execution—essential.


As CAD/CAM-to-construction integration becomes increasingly critical \cite{Intro1, Intro2}, we leverage graph data structures to enhance the assembly sequence of bricks in complex architectural designs, optimizing robotic assembly via sequencing, navigation, and planning \cite{RoboticAssembly1, RoboticAssembly2, Intro1, RoboticAssembly3, RoboticAssembly4, RoboticAssembly5, Intro4, RoboticAssembly8}. 
However, many robotic brick assembly systems still fail to transition beyond laboratory settings \cite{Intro3, Intro4, Intro5}, and traditional learning pipelines (e.g., pre-training, fine-tuning, and transfer learning) struggle with real-world geometric variability \cite{Intro3, Intro4, Intro5}. While \cite{RoboticAssembly7} offers a physics-based planning framework for complex industrial assemblies, it lacks the on-site adaptability required in construction; unlike multi-robot coordination approaches \cite{RoboticAssembly6}, we emphasize single-robot sequencing efficiency to simplify deployment.
Recent planners~\cite{RoboticAssembly7,Fabrica2025,BloxNet2025,PromptToProduct2025} either fix the object class or require per-design optimization, leaving \emph{cross-geometry transfer at deployment} as the unsolved bottleneck for architectural mass customization. This raises a central question: \emph{``Can one planner adapt to unseen topologies in a single forward pass rather than days of retraining, while jointly respecting structural physics and robot kinematics?''}

Current planners face two bottlenecks: lack of universality and operational inefficiency. GRL methods such as GAT-DQN~\cite{Lai2025Poster} and DASTGCN-SNN+DDPG~\cite{Lai2025TechComm} exhibit a "\textbf{retraining bottleneck}"---overfitting to specific topologies and requiring gradient-based retraining when shifting from walls to domes; moreover, sequencing and kinematic execution are commonly decoupled, so stable sequences induce awkward high-energy configurations. To overcome BFS/DFS/A*/GBFS limitations in structural integrity, we employ GRL~\cite{GNN,GCN,GRL,GRL1,GRL2,GRL3,GRL4,GRL5,GRL6} with human priors in 3D modeling. Recent multi-arm GRL~\cite{alt} remains training-heavy, requiring per-structure retraining and relying on sparse end-only rewards with motion planning as a separate bottleneck. This leaves an open gap in \textit{transferable} and \textit{sample-efficient} assembly planning across diverse CAD geometries.

To address these limitations, we present \textit{EUPHORIA}, a framework for universal few-shot adaptability and dynamic efficiency. We argue that a Discrete Element Model (DEM) should act as a "Physics-Regularized" filter: by preemptively pruning physically infeasible actions during training, the simulator shapes the RL policy to respect structural dynamics without sacrificing exploration efficiency. 
Our approach redefines robotic assembly via three contributions:

\begin{itemize} 
\item \textbf{Meta-Geometric Adaptation via Graph Hypernetworks:} Unlike standard feature-matching methods that require retraining, we propose a graph hypernetwork that generates task-specific policy parameters from minimal support sets. This enables unified few-shot planning for unseen topologies (e.g., domes, arches) via parameter-level adaptation.

\item \textbf{Physics-Informed SAC with Kinematics Awareness:} We introduce a \textit{Physics-Bias Attention mechanism} that injects DEM contact forces to prioritize structurally critical connections. Integrated with \textit{Kinematics-Aware Sequencing}, SAC objective optimizes structural stability and \textit{operational efficiency} by penalizing high-energy transitions.

\item \textbf{Differentiable Residual Stability Correction for Sim2Real Robustness:} To bridge the reality gap, we implement a differentiable \textit{Residual Stability Correction} layer. This refines coarse policy actions by minimizing a joint energy–stability cost prior to execution, ensuring robustness against real-world actuation noise and modeling errors.
\end{itemize}

\section{Related Work}

\textbf{Robotic Assembly Planning.} Robotic assembly in construction has been an active area of research, focusing primarily on automating repetitive and labor-intensive tasks \cite{RoboticAssembly1, RoboticAssembly2}. Studies in this domain have explored various methodologies, from basic robotic arms to more complex systems, designed to handle diverse materials and construction scenarios \cite{Intro1, RoboticAssembly3, RoboticAssembly4}. Significant research has been dedicated to robotic bricklaying, with systems like Hadrian X and SAM100 demonstrating practical applications\cite{RoboticAssembly5}, yet these often lack the adaptability required for complex architectural designs \cite{Intro4}. These systems embody advancements in hardware but often rely on predetermined sequences and perceive the environment by vision-based multi-view method\cite{RoboticAssembly8}, limiting their application to designs that deviate from standard layouts. 
Robotic assembly planning has traditionally been dominated by search-based paradigms, including heuristic graph search (e.g., A*) and sampling/search over action trees (e.g., Monte Carlo Tree Search, MCTS). These methods provide strong completeness or bounded-suboptimality guarantees under well-specified cost models, but they often become computationally expensive as the combinatorial space of assembly sequences grows, and they can be brittle when physical constraints, uncertainty, and site-level disturbances must be handled online. In contrast, learning-based approaches aim to amortize planning cost by learning policies from interaction, enabling real-time decision making and adaptation to varying environments. Recent works have also explored graph reinforcement learning (GRL) to exploit relational structure in assembly graphs, enabling policies to reason over brick-to-brick dependencies and structural constraints \cite{GNN, GCN, GRL, GRL1, GRL2, GRL3, GRL4, GRL5, GRL6, Lai2025Poster, Lai2025TechComm}. However, learning-based planners can suffer from heavy training requirements and weak transfer across novel geometries. In particular, Wang et al. \cite{alt} adopt GRL with multi-arm collaboration, but their formulation is training-heavy in practice: it typically requires per-structure retraining and relies on sparse, end-only reward signals, while motion planning is treated as a separate bottleneck. These choices leave an open gap in \textit{transferable} and \textit{sample-efficient} assembly planning across diverse CAD geometries.

\textbf{Mobile Manipulator.} The integration of mobility with robotic manipulation represents a significant evolution in robotic systems, allowing for greater flexibility and a broader range of operations \cite{mobile1}. Mobile manipulators combine the capabilities of robotic arms with mobile platforms, enabling them to perform tasks over a larger area and adapt to varied work environments. Research has primarily focused on developing efficient navigation and manipulation strategies to optimize task execution and ensure safety and accuracy in dynamic environments \cite{Intro1, mobile2, mobile3}. These researches underpin our methodology, where the mobile manipulator is critical for implementation in real-world construction settings.

\textbf{Graph Reinforcement Learning.} Traditional graph search algorithms such as Breadth-First Search (BFS), Depth-First Search (DFS), A* Search, and Greedy Best First Search (GBFS) often struggle to preserve structural integrity in complex assembly configurations, motivating learning-based graph reasoning. Our approach leverages Graph Neural Networks (GNNs) \cite{GNN} to capture dependencies in graph-structured construction data; in particular, Graph Convolutional Networks (GCNs) \cite{GCN} perform convolution directly on graphs to learn node representations that encode neighborhood structure. Graph Reinforcement Learning (GRL) \cite{GRL} further enables decision-making in environments with complex interdependencies \cite{GRL1, GRL2}, and has proven effective across domains where system dynamics are naturally graph-based \cite{GRL3}. Prior work demonstrates graph-based RL for assembly by modeling structural dependencies: Hamrick et al. \cite{GRL4} introduced relational inductive biases for generalization in physical construction, Chung et al. \cite{GRL5} used GNNs to represent geometric structures and optimize assembly sequences, and Ghasemipour et al. \cite{GRL6} scaled structured RL with attention for efficiency and stability. However, most construction/assembly GRL pipelines still rely on fixed-parameter message passing backbones (e.g., GCN/GAT/ASTGCN) specialized to narrow topology families \cite{Lai2025Poster, Lai2025TechComm}, which limits adaptability under substantial geometry/topology shifts and often motivates retraining or extensive fine-tuning.

\textbf{Differentiable Physics \& Control in Robotic Manipulation.}
A complementary thread of research integrates differentiable physics and control to improve physical feasibility and execution quality.
Recent work has advanced differentiable simulation and contact modeling for robotics, enabling gradients through collision/friction events and accelerating downstream optimization and learning \cite{LeLidec2024HEDS, Montaut2023DiffCol, DTAMP}.
In parallel, differentiable tactile/soft-contact simulation has made it increasingly practical to incorporate dense contact feedback into learning and optimization for contact-rich manipulation \cite{Si2024DiffTactile}.
Differentiable visual simulation has also been explored to provide end-to-end gradients for visuomotor policy learning under physical interactions \cite{Huang2023DiffVL}.
Across many pipelines, such optimization is still applied as a post-processing step after a high-level planner produces a discrete plan, potentially introducing latency and decoupling the learning objective from execution quality.
This motivates integrating differentiable optimization as a residual correction layer that refines coarse actions under an explicit energy--stability objective, rather than treating physics-based refinement as an external add-on.


\section{Preliminaries}

\subsection{Problem Formulation}
We formalize the \emph{Simulator-First} automated assembly problem as a tuple $\mathcal{P} = \langle \mathcal{G}_{\text{target}}, \mathcal{R}_{\text{robot}}, \Phi_{\text{DEM}} \rangle$. Here, $\mathcal{G}_{\text{target}} = (\mathcal{V}^*, \mathcal{E}^*)$ represents the target CAD assembly graph, $\mathcal{R}_{\text{robot}}$ denotes the mobile manipulator's configuration space (joints $\mathbf{q} \in \mathbb{R}^n$), and $\Phi_{\text{DEM}}$ is the discrete element model oracle computing contact forces.
The assembly process is modeled as a Finite-Horizon Markov Decision Process (MDP) defined by $(\mathcal{S}, \mathcal{A}, \mathcal{T}, R, \gamma)$, subject to physical constraints.

\paragraph{State Space $\mathcal{S}$.}
At time step $t$, the state $s_t \in \mathcal{S}$ encapsulates the partial assembly and the robot's kinematic state:
\begin{align}
    s_t = \{ \mathcal{G}_t, \mathbf{q}_t \}, \quad \text{with} \quad \mathcal{G}_t = (\mathcal{V}_t, \mathcal{E}_t, \mathbf{X}_t, \mathbf{F}_t).
\end{align}
Here, $\mathcal{G}_t$ is the current brick graph where $\mathcal{V}_t \subseteq \mathcal{V}^*$. Node attributes $\mathbf{X}_t$ contain the position $\mathbf{p}_i \in \mathbb{R}^3$, flattened rotation $\mathbf{r}_i \in \mathbb{R}^9$, and vertex data $V_i$ for each brick $i$. Edge attributes $\mathbf{F}_t$ contain DEM-derived contact forces $F_{ij}$ obtained via $\Phi_{\text{DEM}}(\mathcal{G}_t)$.

\paragraph{Hybrid Action Space $\mathcal{A}$.}
To bridge high-level sequencing with low-level control, we define a hybrid action space $\mathcal{A} = \mathcal{A}_{\text{seq}} \times \mathcal{A}_{\text{res}}$. An action $a_t = (v_k, \delta_t)$ consists of:
\begin{enumerate}
    \item \textbf{Sequence Selection ($v_k$):} A discrete choice $v_k \in \mathcal{V}^* \setminus \mathcal{V}_t$ selecting the next brick to place.
    \item \textbf{Residual Refinement ($\delta_t$):} A continuous vector $\delta_t \in \mathbb{R}^6$ representing a perturbation to the nominal CAD pose $\mathbf{p}_{\text{nom}}(v_k)$. This compensates for Sim2Real discrepancies and optimizes kinematic stability.
\end{enumerate}
The final executed action is $T_{\text{place}} = T_{\text{nom}}(v_k) \oplus \delta_t$.

\subsection{Optimization Objectives and Constraints}
We seek an optimal policy $\pi^*$ that maximizes the expected cumulative reward subject to strict physical feasibility. The optimization problem is defined as:


\vspace{-0.5em}
\begin{equation}
\begin{aligned}
\max_{\pi}\; & J(\pi) = \mathbb{E}_{\tau \sim \pi} \left[ \sum_{t=0}^{T} \gamma^t R(s_t, a_t) \right], \\
\text{s.t.}\; & \textbf{(C1) Stability:} \; \max_{i,j} \| \Phi_{\text{DEM}}(\mathcal{G}_t)_{ij} \| \leq \tau_{\text{fail}}, \\
& \textbf{(C2) Kinematics:} \; \exists\, \xi(t) \in \mathcal{C}_{\text{free}} : \mathbf{q}_{t} \xrightarrow{\xi} \mathbf{q}_{\text{goal}}(a_t), \\
& \textbf{(C3) Geometry:} \; \| \mathcal{G}_T - \mathcal{G}_{\text{target}} \| \to 0.
\end{aligned}
\end{equation}
\vspace{-1.0em}

\begin{figure}
    \centering
    \includegraphics[width=0.5\textwidth]{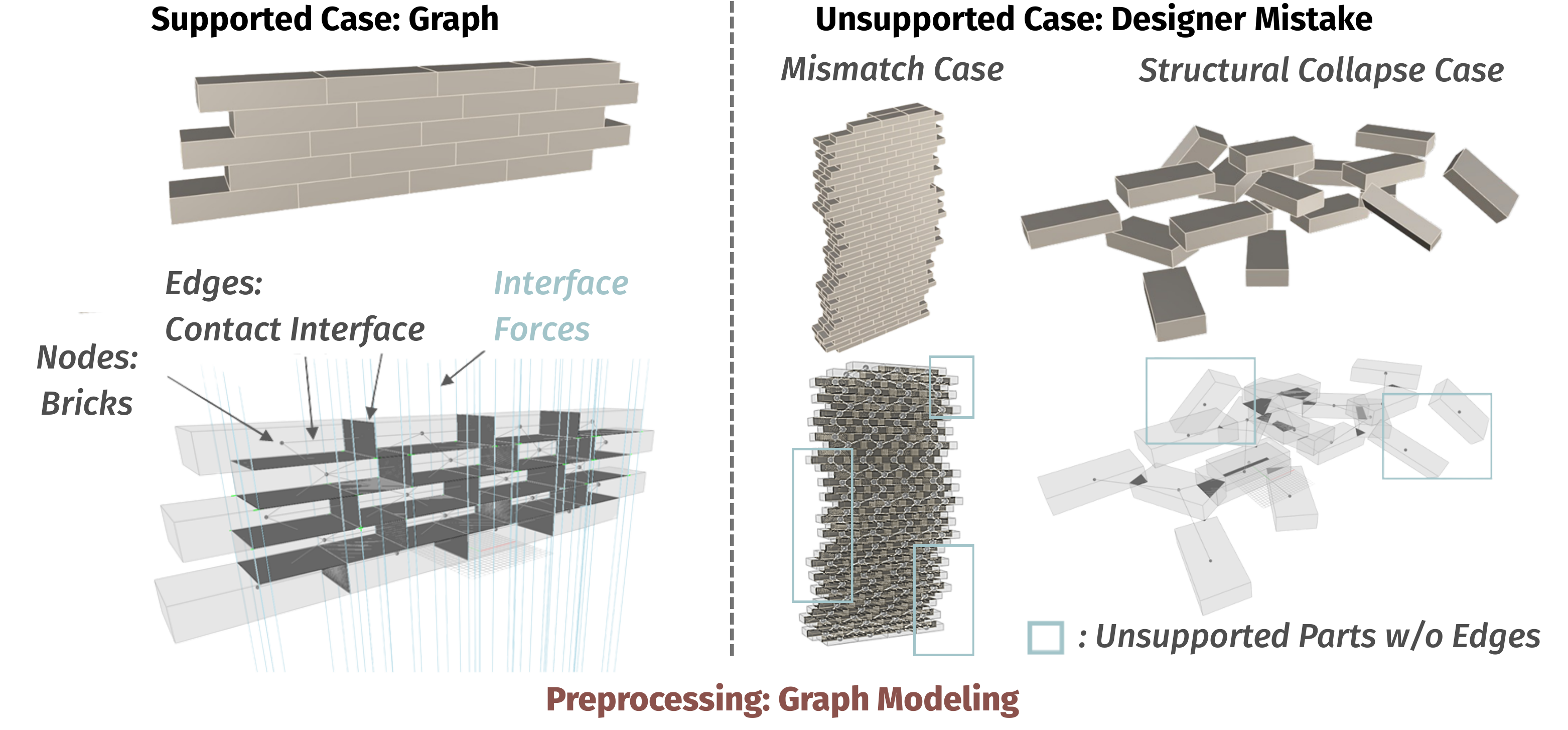} 
    \caption{The bricks' graph modeling in supported case, unsupported (designer mistake: mismatch CAD modeling and collapse case) structure. The deeper gray represents the interface of arbitrary two bricks.}
    \label{Fig_3}
\end{figure}

\section{Method}
\vspace{-0.3em}

\begin{figure*}[t]
  \centering
  \vspace{-0.5em}
  \includegraphics[width=\textwidth]{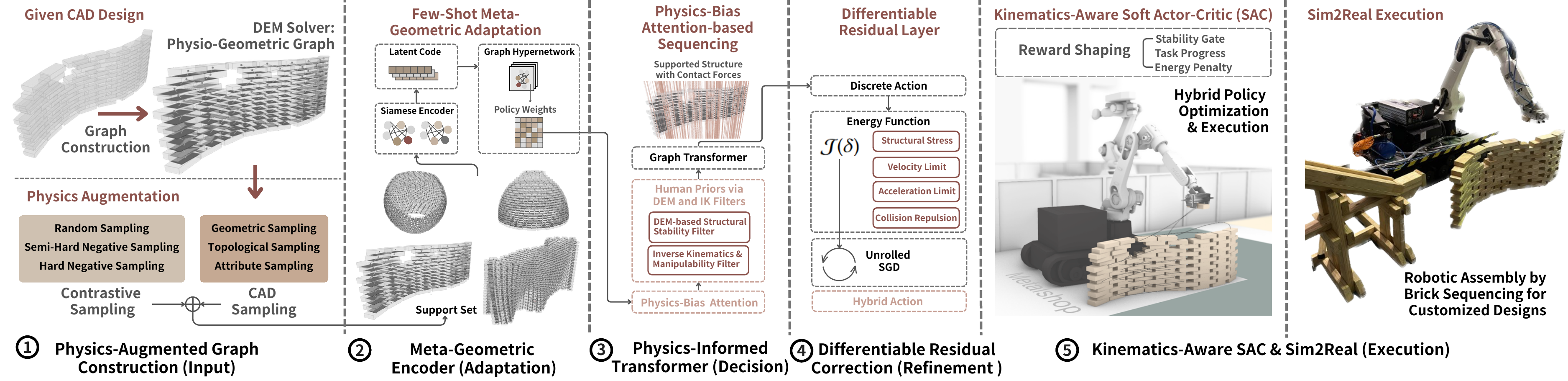}
  \vspace{-0.8em}
  \caption{\textbf{EUPHORIA architecture.} To overcome \textit{retraining for customized designs} and the \textit{operational inefficiency} of  brick sequencing and motion planning, we unify CAD/CAM with robotic assembly via five stages: (1)~\textbf{Physics-Augmented Graph Construction} converts a CAD design into a Physio-Geometric graph via a DEM solver with contrastive/CAD sampling; (2)~a \textbf{Meta-Geometric Encoder} (Siamese + Graph Hypernetwork) generates task-specific policy weights from a few-shot support set; (3)~a \textbf{Physics-Informed Graph Transformer} sequences actions via DEM-biased attention under Human Priors (DEM/IK filters); (4)~a \textbf{Differentiable Residual Layer} refines actions through unrolled SGD on $\mathcal{J}(\delta)$ (stress, velocity, acceleration, collision); (5)~\textbf{Kinematics-Aware SAC} executes the hybrid policy under reward shaping (stability gate, task progress, energy penalty), enabling Sim2Real robotic assembly on unseen geometries.}
  \vspace{-1.0em}
  \label{Fig_1}
\end{figure*}

\subsection{Overview}

\textbf{System Design.}
EUPHORIA is a unified \emph{Simulator-First} framework targeting (i) universal adaptation across CAD geometries and (ii) operational efficiency under robot kinematics. As outlined in \textbf{Algorithm~\ref{alg:euphoria}}, the system builds a brick-level graph from CAD augmented with DEM contact forces (\textit{Fig.~\ref{Fig_1}}); for each task (CAD design) $\tau$, it forms graph state $s_t$ and learns a policy outputting coarse action $a_{\text{coarse}}$, then refines it via a differentiable residual layer for Sim2Real. Unlike fixed-parameter GAT-DQN, EUPHORIA combines (i) a task-conditioned \emph{Meta-Geometric Encoder} for support-set parameter generation, (ii) a \emph{Physics-Informed Graph Transformer} trained via SAC with \emph{Physics-Bias Attention}, (iii) \emph{Kinematics-Aware Sequencing} via energy-penalized reward, and (iv) \emph{Residual Stability Correction} for millimeter-level adjustment (\textit{Fig.~\ref{Fig_3}}).

\noindent\textbf{Hybrid Strategy.} EUPHORIA unifies robotic planning through a three-dimensional hybrid approach: it couples discrete sequencing with continuous residual refinement (\emph{Hybrid Action}), interleaves global Transformer-based planning with local gradient-based correction (\emph{Hybrid Execution}), and jointly optimizes the policy via SAC and the residual layer using differentiable unrolled Stochastic Gradient Descent (SGD)(\emph{Hybrid Optimization}).

\subsection{Data Preprocessing and Sampling}
\label{sec:Data}

\textbf{Dataset.} Each given design type of CAD model $C$ is converted to a brick-level graph sequence for simulator-first assembly. We extract brick geometry (mesh-to-brick instances), compute node attributes $(\mathbf{p}_i, V_i, \mathbf{r}_i)$, define edges $\mathcal{E}$ by adjacency/contact candidates, and attach DEM interface forces by precomputing $F_{ij}$ between bricks using a DEM solver:
\begin{align}
\mathcal{G} = (\mathcal{V}, \mathcal{E}, \mathbf{X}, \mathbf{F}),\qquad
\mathbf{X}=[\mathbf{p}_i, V_i, \mathbf{r}_i]_{i\in\mathcal{V}},\quad
\mathbf{F}=[F_{ij}]_{(i,j)\in\mathcal{E}}.
\end{align}
Here, $\mathbf{p}_i \in \mathbb{R}^3$ denotes the brick position, and $\mathbf{r}_i$ denotes the flattened rotation matrix (replacing $\boldsymbol{\theta}_i$ to avoid confusion with policy parameters). The dataset is $\mathbb{D}=\{\mathcal{G}_\tau\}$ over multiple CAD designs $\tau$.

\textbf{Support Set Construction.} For each task $\tau$, we sample a support set $S=\{g^{(k)}\}_{k=1}^{K}$ with $K=5$ by extracting representative local subgraphs (e.g., arches, junctions, openings) from $\mathcal{G}_\tau$ of given type $C$. These $K$ examples are fed to the Siamese Network \cite{FSL1} to compute the task descriptor $z$, which conditions the Graph Hypernetwork $\mathcal{H}_\phi$ \cite{GH}.

\textbf{Contrastive and CAD Sampling.} To train the Physio-Geometric Embedding to capture ``structural logic,'' we build pairs for contrastive learning from $\mathbb{D}$:
\begin{itemize}
  \item \textit{Random Sampling}: $\{g_i\}_{i=1}^n \overset{\mathrm{iid}}{\sim} \mathbb{D}$.
  \item \textit{Semi-Hard Negative Sampling}: $\{(g_i, g_j)\mid d(f(g_i), f(g_j))<\delta_{\text{semi}},\ g_i\in\mathcal{X}_{\text{pos}},\ g_j\in\mathcal{X}_{\text{neg}}\}$.
  \item \textit{Hard Negative Sampling}: $\{(g_i, g_j)\mid d(f(g_i), f(g_j))<\delta_{\text{hard}},\ g_i\in\mathcal{X}_{\text{pos}},\ g_j\in\mathcal{X}_{\text{neg}}\}$.
\end{itemize}
Here $d(\cdot,\cdot)$ is the embedding distance in the latent space, $f(\cdot)$ is the encoder, and $\mathcal{X}_{\text{pos}}/\mathcal{X}_{\text{neg}}$ denote positive/negative sets defined by CAD-level similarity.
In addition, \textbf{CAD sampling} selects representative subgraphs from a customized CAD model $C$ (\textit{Fig.~\ref{FS_Fig_2}}) via:
(1) \textit{Geometric Sampling} (curvature/shape-driven regions),
(2) \textit{Topological Sampling} (junctions/holes/arches), and
(3) \textit{Attribute Sampling} (regions with distinctive $F_{ij}$ or stress patterns).
We concatenate samples from contrastive and CAD sampling to form diverse, challenging support splits for few-shot adaptation.

\textbf{Denoising Augmentation.} To improve robustness under Sim2Real noise, we apply graph-level perturbations during preprocessing, including random edge removal on $\mathcal{E}$ (to emulate missing contacts) and small perturbations on $\mathbf{X}$ and $\mathbf{F}$ (to emulate sensing/estimation noise). This produces denoised embeddings $z$ and stabilizes downstream policy learning under sparse and noisy interfaces.

\textbf{RL Rollout Construction.} During training, each episode samples a task $\tau$, forms $S$ to generate policy parameters $\theta=\mathcal{H}_\phi(z)$, and collects transitions $(s_t,a_t,R_t,s_{t+1})$. In this process, $a_t$ is refined by the differentiable residual layer, and rewards follow in SAC. These transitions populate the replay buffer $\mathcal{B}$ for SAC updates.

\subsection{Human Priors via DEM and IK Filters}
To enhance learning efficiency and restrict the policy search to the manifold of physically plausible states, we integrate expert knowledge as \emph{hard constraints} acting on the action space. We formalize these priors not as heuristic rewards, but as binary validation functions that prune the feasible set $\mathcal{A}_{\text{feasible}}(s_t)$.

\textbf{DEM-based Structural Stability Filter.}
We rely on the Discrete Element Model oracle $\Phi_{\text{DEM}}$ to evaluate the static equilibrium of a candidate partial assembly. For a candidate action $a$ resulting in a new state state $s'$, the structure is deemed stable if and only if the net force and net moment on every placed brick $i \in \mathcal{V}'$ satisfy the equilibrium conditions within a relaxation tolerance $\epsilon_{\text{tol}}$:
{\small
\begin{align}
    \mathcal{C}_{\text{stable}}(s') = \left\{ s' \;\middle|\; \forall i \in \mathcal{V}', \; 
    \left\| \sum_{j \in \mathcal{N}_i} \mathbf{f}_{ij} + m_i \mathbf{g} \right\| \le \epsilon_{F} 
    \land 
    \left\| \sum_{j \in \mathcal{N}_i} (\mathbf{r}_{ij} \times \mathbf{f}_{ij}) \right\| \le \epsilon_{M} 
    \right\},
\label{eq:dem_stability}
\end{align}
}

where $\mathbf{f}_{ij}$ is the contact force vector derived from $\mathbf{F}' = \Phi_{\text{DEM}}(s')$, $m_i \mathbf{g}$ is the gravitational force, and $\mathbf{r}_{ij}$ is the lever arm from the brick center to the contact point. States violating $\mathcal{C}_{\text{stable}}$ (i.e., immediate collapse or large unbalanced forces) are pruned.

\textbf{Inverse Kinematics (IK) and Manipulability Filter.}
To mitigate singularity-induced failures, we enforce a strict reachability constraint. For a target placement pose $\mathbf{p}_{\text{target}} \in \mathrm{SE}(3)$, a valid configuration must exist in the robot's collision-free configuration space $\mathcal{Q}_{\text{free}}$, and it must maintain sufficient manipulability:
\begin{align}
    \mathcal{C}_{\text{kinematic}}(\mathbf{p}_{\text{target}}) = \left\{ \mathbf{p} \;\middle|\; \exists \mathbf{q} \in \mathcal{Q}_{\text{free}} : \text{FK}(\mathbf{q}) = \mathbf{p} \land w(\mathbf{q}) > \epsilon_{\text{sing}} \right\},
\end{align}
where $w(\mathbf{q}) = \sqrt{\det(\mathbf{J}(\mathbf{q})\mathbf{J}(\mathbf{q})^\top)}$ is the Yoshikawa manipulability index. This ensures the robot operates away from singular configurations where $\det(\mathbf{J}) \to 0$, preventing velocity spikes and control instability.

\textbf{Filtered Graph Construction.}
Incorporating these priors, the effective assembly graph state at time $t$ is formally defined as:
\(
    \mathcal{G}_t = (\mathcal{V}_t, \mathcal{E}_t, \mathbf{X}_t, \mathbf{F}_t).
\)
Crucially, the edge set $\mathcal{E}_t$ is not merely geometric adjacency, but the intersection of geometric, structural, and kinematic validity:
\begin{align}
    \mathcal{E}_t = \left\{ (i,j) \in \mathcal{V}_t \times \mathcal{V}_t \;\middle|\; \text{dist}(i,j) < \delta \land s_t \in \mathcal{C}_{\text{stable}} \land \mathbf{p}_i \in \mathcal{C}_{\text{kinematic}} \right\}.
    \label{eq:filtered_edges}
\end{align}
By defining edges through these strict constraints, we ensure that the Graph Transformer propagates information only through physically and kinematically valid load paths. This explicit filtering significantly reduces the search space, as demonstrated in our \textbf{Ablation Study}, where removing these priors results in a marked degradation in convergence speed and assembly success rate.

\subsection{Planning for Graph Sequencing-Based Brick Assembly}
We formulate the assembly process as a Partially Observable Markov Decision Process (POMDP) on graph-structured states, defined by the tuple $(\mathcal{S}, \mathcal{A}, \mathcal{P}, \mathcal{R}, \gamma)$. At each time step $t$, the agent observes the graph state $s_t$ derived from the partial assembly $\mathcal{G}_t$ and selects a composite action $a_t = (a_{\text{seq}}, \delta_t)$, where $a_{\text{seq}}$ is the discrete sequencing decision and $\delta_t$ is the continuous residual refinement. To achieve universal adaptation and dynamic efficiency, EUPHORIA employs a hierarchical differentiable architecture comprising four mathematically rigorous modules.


\begin{figure}[t]
\centering
  \includegraphics[width=0.8\linewidth]{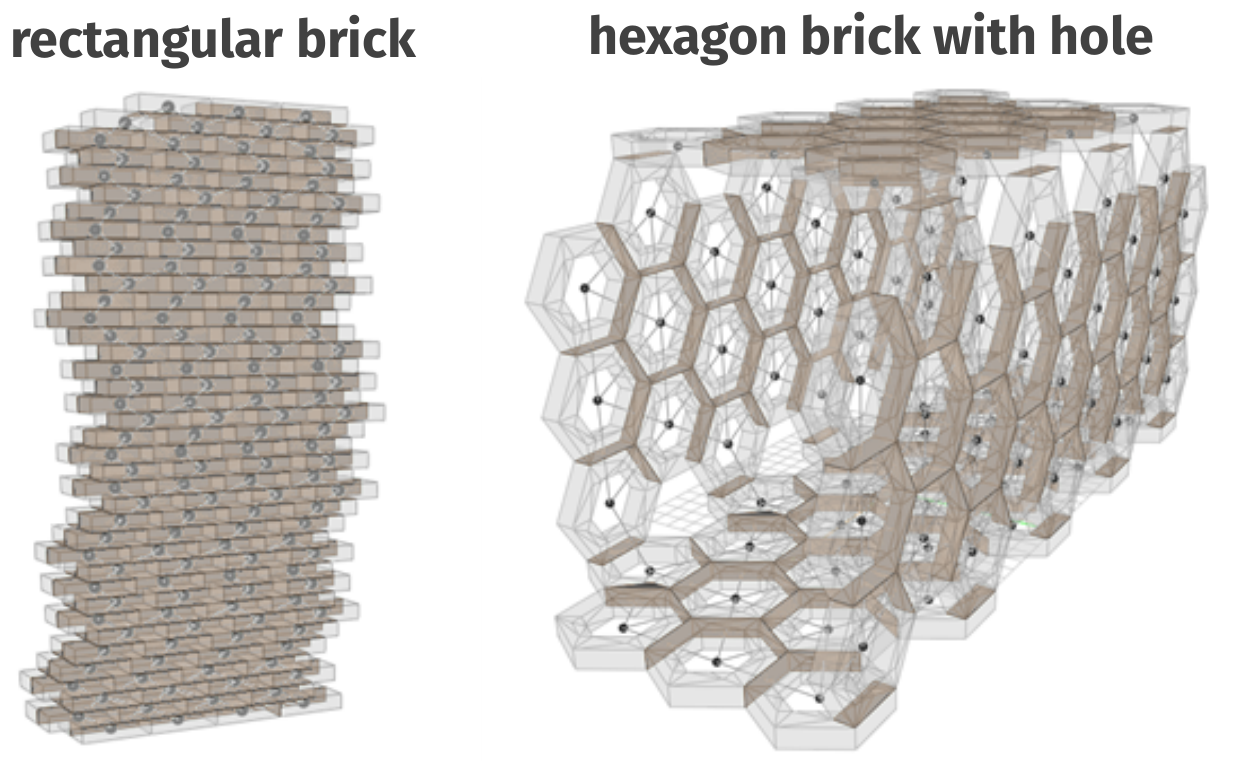}
  \caption{\textbf{Topological Generalization.} Visual comparison between \textbf{(Left)} structured rectangular lattices used in training and \textbf{(Right)} unstructured hexagonal geometries used for few-shot adaptation evaluation. 
  }
  \label{Fig_Topology_Shift}
\end{figure}

\begin{algorithm}[h]
\vspace{-0.3em}
\caption{EUPHORIA}
\label{alg:euphoria}
\begin{algorithmic}[1]
\Require Task support set $S=\{g^{(k)}\}_{k=1}^K$; DEM Oracle $\Phi_{\text{DEM}}$.

\State \textbf{1. Few-Shot Policy Meta-Geometric Adaptation:}
\State Extract task descriptor $z \leftarrow f_\psi(S)$ \Comment{Encode $K$ examples}
\State Generate task-specific planner parameters $\theta_\tau \leftarrow \mathcal{H}_\phi(z)$ \Comment{Via Graph Hypernetwork, Eq.~\ref{eq:hyper_generation}; Adapt weights without retraining}

\State \textbf{2. Hybrid Execution Loop:}
\For{$t = 0, \dots, T-1$}
    \State \textit{// Physics-Informed SAC w/ Kinematics (Discrete Sequencing)}
    \State Compute DEM contact forces $\mathbf{F}_t \leftarrow \Phi_{\text{DEM}}(\mathcal{G}_t)$
    \State Compute attention $e_{ij}$ biased by $\log(1+\|\mathbf{F}_t\|)$ \Comment{Eq.~\ref{eq:physics_attention}}
    \State Sample coarse action $a_{\text{coarse}} \sim \pi_{\theta_\tau}(s_t)$
    
    \State \textit{// Differentiable Residual Correction (Continuous Refinement)}
    \State Solve $\delta_t^* \leftarrow \text{UnrolledSGD}(\mathcal{J}, a_{\text{coarse}})$ \Comment{Sim2Real Refinement, Eq.~\ref{eq:residual_energy}}
    \State Execute Hybrid Action $a_t \leftarrow a_{\text{coarse}} + \delta_t^*$
    
    \State \textit{// Environment Feedback}
    \State Observe $R_t$ with stability gate and energy penalty \Comment{Eq.~\ref{eq:reward_master}}
    \State Store transition in $\mathcal{B}$
\EndFor

\State \textbf{3. Optimization:}
\State Update Critic $\omega$  via Bellman error and Hypernetwork $\phi$ using SAC loss backpropagated through Unrolled SGD steps.
\State Export assembly sequence of bricks $a_t = (a_{\text{seq}}, \delta_t)$
\end{algorithmic}
\end{algorithm}

\subsubsection{Few-Shot Meta-Geometric Encoder \& Hypernetwork}
\label{sec:few_shot_encoder}

Standard policy networks with fixed weights $\theta$ fail to generalize to unseen topological structures (e.g., transitioning from walls to domes). We resolve this via a \textit{Graph Hypernetwork} that enables \emph{few-shot parameter-level adaptation}. A critical requirement is the policy's ability to generalize to unseen topologies $\mathcal{G}_{\text{new}}$ without gradient-based retraining. We formalize this as a \emph{parameter generation} problem.

\textbf{Few-Shot Support Set.}
For a new task $\tau$, we construct a small support set $S = \{g^{(k)}\}_{k=1}^K$ containing $K=5$ representative subgraphs (e.g., arches). This set $S$ serves as the few-shot signal.

\textbf{Physio-Geometric Embedding.}
We employ a Siamese Graph Encoder $f_{\psi}$ to map the few-shot examples $S$ into a single latent task descriptor $z_\tau$ for encoding the structural logic. The encoder is trained via a Triplet Loss $\mathcal{L}_{\text{triplet}}$ to ensure that geometrically distinct but structurally functionally similar substructures are clustered in the latent space $\mathcal{Z}$:
\(
    z_\tau = \frac{1}{K} \sum_{k=1}^{K} f_{\psi}(g^{(k)}), \quad \text{where } z_\tau \in \mathbb{R}^{d_z}.
\)

\textbf{Dynamic Weight Generation via Graph Hypernetworks.}
To enable universal adaptation (\textit{Fig.~\ref{Fig_Topology_Shift}}), we seek a hyper-function  that dynamically maps the latent structural descriptor  to the task-specific policy parameters. Specifically, for each layer  of the downstream planner, the Graph Hypernetwork generates the specific weight matrices  and biases  via:
{\small
\begin{align}
\theta_{\tau}^{(l)} = \{\mathbf{W}^{(l)}, \mathbf{b}^{(l)}\} = \mathrm{Reshape}\!\left(\mathrm{MLP}_{\phi}^{(l)}\!\left(\frac{1}{K}\sum_{k=1}^{K} f_{\text{enc}}(g^{(k)})\right)\right).
\label{eq:hyper_generation}
\end{align}
}
This mechanism allows the policy  to instantiate a task-specific manifold in the parameter space, effectively "morphing" the planner into a specialist for the new CAD geometry in a single inference pass. By conditioning the weights directly on the few-shot support set, the system adapts to the specific structural logic of unseen designs without the need for gradient-based retraining.

\subsubsection{Physics-Informed Graph Transformer (PIGT)}

The generated policy $\pi_{\theta_\tau}$ is instantiated as a Physics-Informed Graph Transformer. Unlike standard Graph Attention Networks (GATs) that rely solely on geometric adjacency, PIGT integrates a \textit{Physics-Bias Attention (PBA) Mechanism (Force-Guided Attention)} to explicitly model force transmission paths, leveraging the Hypernetwork weights $\theta_\tau = \{ \mathbf{W}_Q, \mathbf{W}_K, \mathbf{W}_V, \dots \}$ derived in the previous step. Let $\mathbf{H}^{(l)} \in \mathbb{R}^{N \times d_h}$ be the node feature matrix at layer $l$, where $\mathbf{H}^{(0)}$ corresponds to the node attributes $\mathbf{X}$ defined in Sec. \ref{sec:Data}. The query, key, and value are computed using the task-specific weights generated by the Hypernetwork: 
\(
    \mathbf{Q} = \mathbf{H}^{(l)}\mathbf{W}_Q^{(l)}, \quad \mathbf{K} = \mathbf{H}^{(l)}\mathbf{W}_K^{(l)}, \quad \mathbf{V} = \mathbf{H}^{(l)}\mathbf{W}_V^{(l)}.
\)

\textbf{Physics-Bias Attention Score.}
Standard attention mechanisms treat all geometric neighbors equally. To incorporate structural logic, we introduce a structural coupling bias derived from the precomputed DEM contact forces $F_{ij}$. The attention coefficient $e_{ij}$ between brick $i$ and brick $j$ is formulated as:
{\small
\begin{align}
    e_{ij} = \frac{\mathbf{q}_i^\top \mathbf{k}_j}{\sqrt{d_k}} + \underbrace{\beta \cdot \log(1 + \|F_{ij}\|_2)}_{\text{Physics Bias}} + \mathcal{M}_{ij},
    \label{eq:physics_attention}
\end{align}
}
where $\mathbf{q}_i, \mathbf{k}_j$ are row vectors of $\mathbf{Q}$ and $\mathbf{K}$, $\beta$ is a learnable scaling factor that calibrates the influence of physics versus geometry, and $\|F_{ij}\|_2$ is the normalized magnitude of the contact force. $\mathcal{M}_{ij}$ is the structural mask enforcing the filtered topology ($\mathcal{M}_{ij}=0$ if $j \in \mathcal{E}_t$, $-\infty$ otherwise). The normalized attention weights $\alpha_{ij}$ and the layer output are:
\begin{align*}
    \alpha_{ij} &= \text{softmax}_j(e_{ij}), \\
    \mathbf{h}_i^{(l+1)} &= \text{LayerNorm}\!\left( \mathbf{h}_i^{(l)} + \sum_{j \in \mathcal{N}_i} \alpha_{ij} \mathbf{v}_j \right).
\end{align*}
This mechanism forces the policy to assign higher importance to structurally critical nodes (e.g., load-bearing joints or keystones) regardless of mere geometric proximity. Finally, the policy outputs a categorical distribution for the coarse sequencing action by projecting the final node embeddings:
\(
    a_{\text{coarse}} \sim \text{Categorical}(\text{MLP}_{\theta_\tau}(\mathbf{h}^{(L)})),
\)
where the MLP weights are conditioned on the task descriptor $z_\tau$.

\subsubsection{Differentiable Residual Stability Correction}
\label{sec:diff_residual}

The coarse sequencing action $a_{\text{coarse}}$ provided by the Transformer is insufficient for Sim2Real due to discretization errors. We refine it via a continuous residual vector $\delta_t \in \mathbb{R}^6$, modeled as a differentiable optimization layer. We define a unified potential function $\mathcal{J}(\delta)$ that balances structural stability, kinematic feasibility, and collision avoidance:
{\small
\vspace{-0.5em}
\begin{equation}
\label{eq:residual_energy}
\begin{split}
    \mathcal{J}(\delta) &= \underbrace{w_s \|\sigma_{\text{surrogate}}(a_{\text{coarse}} + \delta)\|^2_F}_{\text{Structural Stress}} 
    + \underbrace{\lambda_1 \|\mathbf{J}(\mathbf{q}(\delta))\,\dot{\mathbf{q}}(\delta)\|^2}_{\text{Velocity Limit}} \\
    &\quad + \underbrace{\lambda_2 \|\ddot{\mathbf{q}}(\delta)\|^2}_{\text{Acceleration Limit}} 
    + \underbrace{\sum_{o=1}^{M} k_{\text{col}} \exp\!\left(-\frac{\|\mathrm{FK}(\mathbf{q}(\delta)) - \mathbf{x}_{\text{obs}, o}\|^2}{2\sigma^2}\right)}_{\text{Collision Repulsion (task space)}}.
\end{split}
\end{equation}
\vspace{-1.0em}
}

All kinematic quantities are functions of $\delta$ through the residual placement map $\mathbf{p}(\delta) = \mathbf{p}_{\text{nom}}(v_k) \oplus \delta$ followed by inverse kinematics: $\mathbf{q}(\delta) = \mathrm{IK}(\mathbf{p}(\delta))$, with $\dot{\mathbf{q}}$ and $\ddot{\mathbf{q}}$ obtained by finite differences along the executed trajectory. Here, $\sigma_{\text{surrogate}}$ is a fast differentiable approximation of the stress tensor (modeled via a lightweight MLP), $\mathbf{J}(\mathbf{q})$ is the manipulator Jacobian, $\mathrm{FK}: \mathbb{R}^n \to \mathrm{SE}(3)$ denotes the forward-kinematics map from joint configuration to end-effector pose, and $\mathbf{x}_{\text{obs}, o} \in \mathbb{R}^3$ is the position of obstacle $o$ (a placed brick or static environment element). The refined action is obtained by solving the unconstrained minimization $\delta_t^* = \arg\min_{\delta} \mathcal{J}(\delta)$. Measuring repulsion in \emph{task space} avoids the redundant joint-space mapping; $\nabla_\delta$ propagates through $\mathrm{FK}$ via the Jacobian $\mathbf{J}(\mathbf{q})$ already computed for the velocity term, requiring no extra differentiation.

\textbf{Differentiability via Unrolled SGD.}
To ensure end-to-end differentiability, we implement the solver as a \textit{Unrolled Stochastic Gradient Descent (SGD)} module embedded within the computation graph. Let $\delta^{(0)} = \mathbf{0}$. For $k=0 \dots K-1$:
{\small
\begin{align}
    \delta^{(k+1)} \leftarrow \delta^{(k)} - \eta \nabla_\delta \mathcal{J}(\delta^{(k)}).
    \label{eq:residual_opt}
\end{align}
}

Since each update consists of differentiable operations, gradients from the downstream RL objective backpropagate through these $K$ steps. This updates the upstream policy to produce coarse actions that are energy-efficient and "easily correctable" by residual layer.

\subsubsection{Kinematics-Aware Soft Actor-Critic (SAC)}

To enable dynamic efficiency, we employ Maximum Entropy RL. We train the policy $\pi_{\theta}$ (whose parameters $\theta$ are generated by $\mathcal{H}_\phi$) and the soft Q-function $Q_\omega$ to maximize the expected return augmented by an entropy term.

\textbf{Reward Formulation.}
To realize \emph{Kinematics-Aware Sequencing}, the reward function $R(s_t, a_t)$ is rigorously defined to enforce a hierarchy of objectives: (1) strict structural stability, (2) task progression, and (3) minimization of kinematic energy.
{\small
\begin{align}
    R(s_t, a_t) = \underbrace{\mathbb{I}(\lambda_{\min}(\mathbf{L}_{t+1}) > 0)}_{\text{Stability Gate}} \cdot \left( R_{\text{task}}(s_{t+1}) - \lambda_{\text{energy}} \cdot C_{\text{kinematic}}(s_t, a_t) \right) + R_{\text{fail}},
    \label{eq:reward_master}
\end{align}
}
where $\mathbb{I}(\cdot)$ is the indicator function derived from the DEM simulator acting as a hard constraint.
The components are defined as:
\begin{itemize}
    \item \textit{Task Progress ($R_{\text{task}}$):}
    \small{
    \begin{align}
        R_{\text{task}} = w_c \log(1 + N_{t+1}) + w_p \exp\left( -\frac{\| \mathbf{x}_{\text{target}} - \mathbf{x}_{\text{placed}} \|^2}{2\sigma^2} \right).
    \end{align}
    }
    \item \textit{Kinematic Cost ($C_{\text{kinematic}}$):}
    \small{
    \begin{align}
        C_{\text{kinematic}} = \int_{t}^{t+1} \boldsymbol{\tau}^\top \dot{\mathbf{q}} \, dt + \rho \cdot \mathbb{I}_{\text{singularity}}(\mathbf{q}),
    \end{align}
    }
    where $\mathbb{I}_{\text{singularity}}$ penalizes configurations near kinematic singularities (manipulability index $<\epsilon$).
\end{itemize}

The network is trained via standard SAC objectives adapted for hypernetworks, maximizing the entropy-regularized expected return (detailed in \textbf{Supplementary Sec.~\ref{OptObj}}).

\section{Baselines}
\label{sec:baselines}

To benchmark EUPHORIA against state-of-the-art approaches, we compare it with a unified suite of baselines across Sequencing, Spatiotemporal Graph (STG), Few-Shot Learning (FSL), and Graph RL (GRL) domains. All baselines share the same preprocessing pipeline and access to  and DEM-derived  for fair comparison.

\begin{itemize}
\item \textbf{Search and Standard RL Baselines:} We compare against classical search methods (\textbf{DFS}, \textbf{BFS}, \textbf{A*}, \textbf{GBFS}) augmented with Stability Checks (SC) using the constraint . Learning baselines include \textbf{DQN+GAT}, \textbf{GCQN}, \textbf{DQN+GNN}, \textbf{GNNPG}, and \textbf{GCPN}, as well as a fixed-parameter \textbf{GCN+RL} baseline (without hypernetwork adaptation).

\item \textbf{Spatiotemporal and Few-Shot Baselines:} To evaluate representation power, we compare with STG models including \textbf{STGCN} \cite{STG1}, \textbf{GAT} \cite{STG2}, \textbf{GraphSAINT} \cite{STG3}, \textbf{Pro-GNN} \cite{STG4}, \textbf{PTDNet} \cite{STG5}, \textbf{DGCRN} \cite{STG6}, \textbf{GMAN} \cite{STG7}, \textbf{ASTGCN} \cite{STG8}, and \textbf{GraphSAGE-GAT} \cite{STG9}. Few-shot generalization capabilities are tested against \textbf{Siamese Neural Networks (SNN)} \cite{FSL1}, \textbf{MAML} \cite{FSL2}, \textbf{RelationNet} \cite{FSL3}, \textbf{LEO} \cite{FSL4}, and \textbf{ProtoNet} \cite{FSL5} under the same support/query protocol.

\item \textbf{Graph Reinforcement Learning (GRL) Baselines:} Decision quality is evaluated against \textbf{A2C}-based methods (\textbf{A2C-GNN} \cite{GRL_1}, \textbf{A2C-GAT} \cite{STG2}, \textbf{A2C-Pro-GNN} \cite{STG4}, \textbf{A2C-PTDNet} \cite{STG5}) and \textbf{DDPG}-based methods (\textbf{DDPG-GNN} \cite{GRL_4}, \textbf{DDPG-GAT} \cite{STG2}, \textbf{DDPG-Pro-GNN} \cite{STG4}, \textbf{DDPG-PTDNet} \cite{STG5}), alongside \textbf{MA-GCN} \cite{GRL_9} and \textbf{GNN-UCB} \cite{GRL_10}. Furthuremore, \textbf{Prior Pipeline (DASTGCN-SNN+DDPG):} combining \textbf{DASTGCN} (with DropEdge \cite{AS2}, GAT \cite{STG2}, Graph Laplacian Regularization \cite{AS1}), an SNN metric learner \cite{FSL1}, and a DDPG policy \cite{GRL_4}. It serves as the primary reference for few-shot adaptation without EUPHORIA's unified hypernetwork and physics-bias mechanisms. We also report ablations for DASTGCN components (``w/o DropEdge'', ``w/o GAT'', ``w/o Reg'') to isolate denoising contributions.

\item \textbf{EUPHORIA Ablations:} To validate our contributions, we test variants: (i) \textbf{w/o Human Priors (HP)}, (ii) \textbf{w/o Physics-Bias Attention}, (iii) \textbf{Decoupled Planning} (sequencing then motion planning), and (iv) \textbf{w/o Residual Stability Correction}.

\end{itemize}

\begin{figure*}[!t]
    \centering 
    \includegraphics[width=\textwidth]{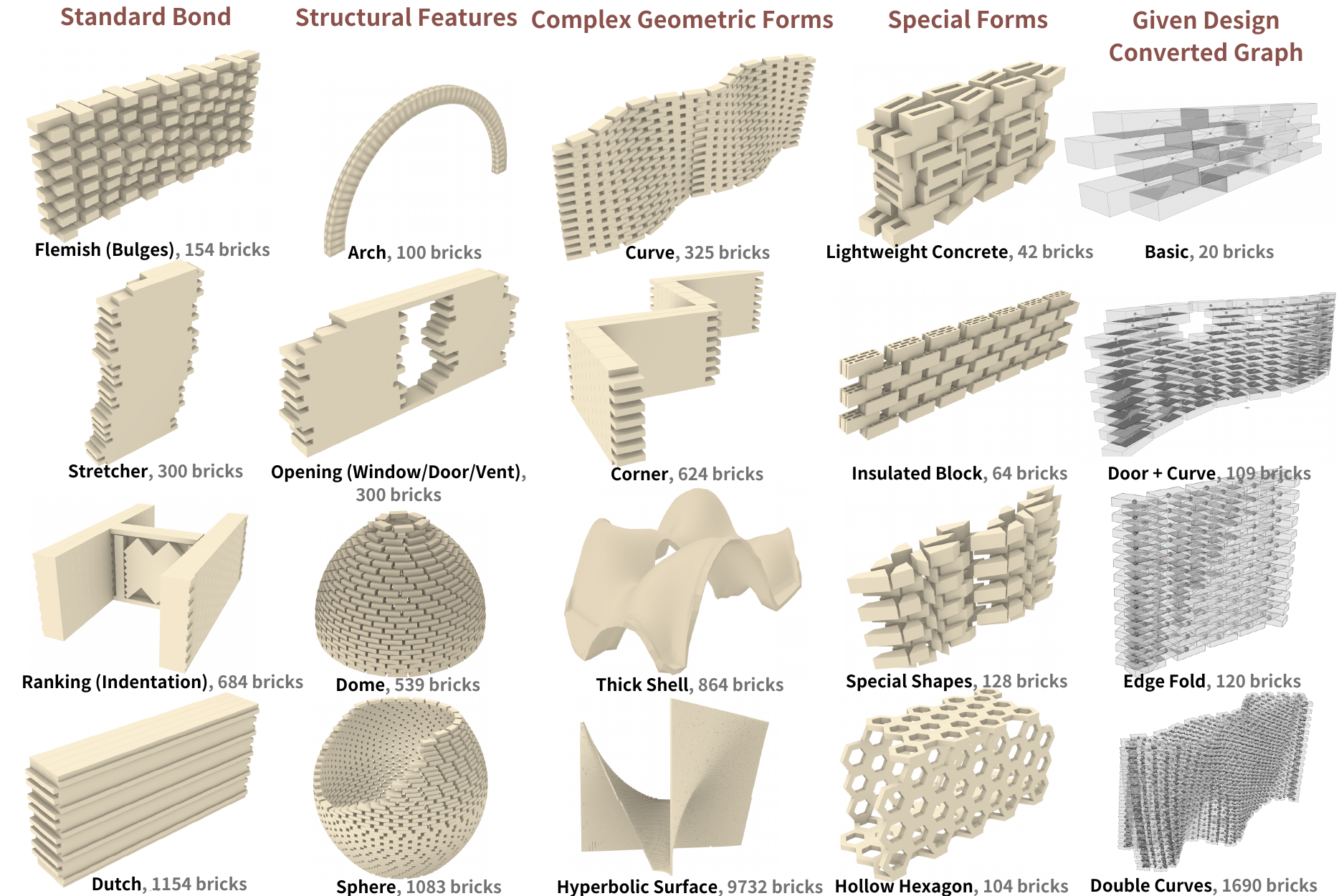}
    \caption{\textbf{Dataset Taxonomy and Graph Representation.} 
    We curate a diverse parametric CAD models categorized into four structural complexity levels: 
    \textit{(1) Standard Bonds} (e.g., Flemish, Dutch layouts), 
    \textit{(2) Structural Features} (e.g., Arches, Domes), 
    \textit{(3) Complex Geometric Forms} (e.g., Hyperbolic Surfaces, Corners), and 
    \textit{(4) Special Forms} (e.g., Hollow Hexagons, Insulated Blocks),
    spanning $N=20$ to $N=9{,}732$ bricks across 50 parametric varied evaluation instances ( curvature, opening size, course count, etc.) to test scalability. The rightmost column shows the \textit{Graph Conversion} process: continuous CAD geometries are abstracted into discrete connectivity graphs $\mathcal{G}$ where transparent nodes $\mathcal{V}$ and contact edges $\mathcal{E}$ enable the planner to check stability and sequencing.}
    \label{FS_Fig_2}
\end{figure*}

\begin{table*}[t]
\centering
\setlength{\abovecaptionskip}{2pt}
\setlength{\belowcaptionskip}{0pt}
\captionsetup{skip=2pt,belowskip=0pt}
\captionsetup[subtable]{skip=1pt,belowskip=0pt}
\caption{\textbf{Baseline Comparison in Simulation-based Training: Representation and Policy Quality.} Comparisons with baselines, EUPHORIA achieves superior performance across all metrics, validating the hybrid optimization framework on a mixed dataset of diverse brick types.}
\label{tab:comparison}
\begin{adjustbox}{width=\textwidth}
\begin{tabular}{@{}p{0.33\textwidth} p{0.33\textwidth} p{0.33\textwidth}@{}}
\centering
\subcaption{Spatiotemporal Graph Learning}
\vspace{-0.2em}
\begin{adjustbox}{width=0.9\linewidth}
\begin{tabular}{@{}lccc@{}}
\toprule
    Model & MSE $\downarrow$ & MAE $\downarrow$ & TCC $\uparrow$ \\
    \midrule
    STGCN \cite{STG1} & 0.040 & 0.15 & 0.82 \\
    GAT \cite{STG2}& 0.035 & 0.14 & 0.85 \\
    GraphSaint \cite{STG3}& 0.032 & 0.13 & 0.83 \\
    Pro-GNN \cite{STG4}& 0.038 & 0.16 & 0.80 \\
    PTDNet \cite{STG5}& 0.030 & 0.15 & 0.81 \\
    DGCRN \cite{STG6}& 0.033 & 0.13 & 0.87 \\
    GMAN \cite{STG7}& 0.034 & 0.14 & 0.86 \\
    ASTGCN \cite{STG8}& 0.031 & 0.12 & 0.89 \\
    GraphSAGE-GAT \cite{STG9}& 0.026 & 0.15 & 0.90 \\
    \textit{GAT-DQN} & 0.028 & 0.12 & 0.88 \\
    \textit{DASTGCN-SNN+DDPG} & \underline{0.025} & \underline{0.10} & \underline{0.92} \\
    \textbf{EUPHORIA (Ours)} & \textbf{0.020} & \textbf{0.08} & \textbf{0.95} \\
\bottomrule
\end{tabular}
\end{adjustbox}
&
\centering
\subcaption{Few-Shot Learning}
\vspace{-0.2em}
\begin{adjustbox}{width=1\linewidth}
\begin{tabular}{@{}lcccc@{}}
\toprule
Model & Acc $\uparrow$ & F1 $\uparrow$ & NMI $\uparrow$ & Loss $\downarrow$ \\
\midrule
SNN \cite{FSL1}& 0.75 & 0.70 & 0.60 & 0.25 \\
MAML \cite{FSL2}& 0.78 & 0.73 & 0.65 & 0.22 \\
RelationNet \cite{FSL3}& 0.77 & 0.72 & 0.68 & 0.23 \\
LEO \cite{FSL4}& 0.79 & 0.74 & 0.67 & 0.20 \\
ProtoNet \cite{FSL5}& 0.80 & 0.75 & 0.70 & 0.24 \\
\textit{GAT-DQN} & 0.82 & 0.76 & 0.72 & 0.21 \\
\textit{DASTGCN-SNN+DDPG} & \underline{0.85} & \underline{0.80} & \textbf{0.75} & \underline{0.18} \\
\textbf{EUPHORIA (Ours)} & \textbf{0.90} & \textbf{0.85} & \underline{0.73} & \textbf{0.15} \\
\bottomrule
\end{tabular}
\end{adjustbox}
&
\centering
\subcaption{Graph Reinforcement Learning}
\vspace{-0.2em}
\begin{adjustbox}{width=0.7\linewidth}
\begin{tabular}{@{}lccc@{}}
\toprule
    Model & $R$ $\uparrow$ & $\overline{r}$ $\uparrow$ & $c$ $\downarrow$ \\
    \midrule
    A2C-GNN \cite{GRL_1} & 150 & 1.5 & 0.39 \\
    A2C-GAT \cite{STG2} & 152 & 1.52 & 0.42 \\
    A2C-Pro-GNN \cite{STG4} & 148 & 1.48 & 0.51 \\
    A2C-PTDNet \cite{STG5} & 149 & 1.49 & 0.44 \\
    DDPG-GNN \cite{GRL_4} & 160 & 1.6 & 0.25 \\
    DDPG-GAT \cite{STG2} & 162 & 1.62 & 0.23 \\
    DDPG-Pro-GNN \cite{STG4} & 158 & 1.58 & 0.27 \\
    DDPG-PTDNet \cite{STG5} & 159 & 1.59 & 0.22 \\
    MA-GCN \cite{GRL_9} & 167 & 1.57 & 0.28 \\
    GNN-UCB \cite{GRL_10} & 155 & 1.65 & 0.23 \\
    \textit{GAT-DQN} & 165 & 1.65 & 0.24 \\
    \textit{DASTGCN-SNN+DDPG} & \underline{170} & \underline{1.70} & \underline{0.20} \\
    \textbf{EUPHORIA (Ours)} & \textbf{185} & \textbf{1.85} & \textbf{0.18} \\
\bottomrule
\end{tabular}
\end{adjustbox}
\end{tabular}
\end{adjustbox}
\end{table*}

\begin{table*}[t]
\centering
\caption{\textbf{Performance in Brick Assembly Sequencing and Planning in Simulation.} 
EUPHORIA is compared against other planning baselines. EUPHORIA's hybrid optimization significantly improves Path Efficiency and Stability over the baselines. (Mean $\pm$ SD over 5 runs).}
\label{tab:model_comparison}
\resizebox{\textwidth}{!}{
\begin{tabular}{>{\bfseries}lcccccccccc}
\toprule
\textbf{Metric} & \textbf{EUPHORIA} & \textbf{DASTGCN-SNN+DDPG} & \textbf{GAT-DQN} & \textbf{GCQN} & \textbf{DQN+GNN} & \textbf{GNNPG} & \textbf{A*+SC} & \textbf{GBFS+SC} & \textbf{DFS+SC} & \textbf{BFS+SC} \\
\midrule
Action Entropy ($H$) $\uparrow$
  & \textbf{0.94$\pm$0.01}
  & \underline{0.93$\pm$0.01}
  & 0.92$\pm$0.01 
  & 0.85$\pm$0.015 
  & 0.87$\pm$0.012 
  & 0.90$\pm$0.01 
  & 0.78$\pm$0.02 
  & 0.80$\pm$0.02 
  & 0.74$\pm$0.025 
  & 0.55$\pm$0.03 \\

Stability Index ($\lambda_{\min}$) $\uparrow$
  & \textbf{0.18$\pm$0.005}
  & \underline{0.16$\pm$0.005}
  & 0.15$\pm$0.005 
  & 0.10$\pm$0.005 
  & 0.12$\pm$0.005 
  & 0.11$\pm$0.005 
  & 0.10$\pm$0.006 
  & 0.11$\pm$0.006 
  & 0.07$\pm$0.007 
  & 0.05$\pm$0.007 \\

Attention Consistency (AC) $\uparrow$
  & \textbf{0.97$\pm$0.01}
  & \underline{0.96$\pm$0.01}
  & 0.95$\pm$0.01 
  & 0.89$\pm$0.015 
  & 0.91$\pm$0.01 
  & 0.88$\pm$0.012 
  & 0.85$\pm$0.02 
  & 0.86$\pm$0.02 
  & 0.73$\pm$0.025 
  & 0.62$\pm$0.03 \\

Temporal Credit Assignment (TCA) $\uparrow$
  & \textbf{0.92$\pm$0.01}
  & \underline{0.90$\pm$0.01}
  & 0.88$\pm$0.015 
  & 0.82$\pm$0.02 
  & 0.85$\pm$0.015 
  & 0.84$\pm$0.017 
  & 0.77$\pm$0.03 
  & 0.79$\pm$0.025 
  & 0.61$\pm$0.03 
  & 0.59$\pm$0.03 \\

Path Efficiency ($P_{\text{eff}}$) $\uparrow$
  & \textbf{0.85$\pm$0.02}
  & \underline{0.80$\pm$0.02}
  & 0.75$\pm$0.03 
  & 0.65$\pm$0.04 
  & 0.70$\pm$0.03 
  & 0.68$\pm$0.035 
  & 0.66$\pm$0.04 
  & 0.69$\pm$0.035 
  & 0.57$\pm$0.05 
  & 0.43$\pm$0.05 \\

Structural Integrity ($S_{\text{integ}}$) $\uparrow$
  & \textbf{0.88$\pm$0.01}
  & \underline{0.84$\pm$0.01}
  & 0.80$\pm$0.02 
  & 0.72$\pm$0.03 
  & 0.75$\pm$0.02 
  & 0.74$\pm$0.025 
  & 0.72$\pm$0.03 
  & 0.73$\pm$0.025 
  & 0.67$\pm$0.04 
  & 0.55$\pm$0.05 \\

Completion Time (CT, second) $\downarrow$
  & \textbf{105$\pm$4}
  & 120$\pm$5
  & \underline{112$\pm$4} 
  & 140$\pm$7 
  & 135$\pm$6 
  & 130$\pm$5 
  & 150$\pm$8 
  & 145$\pm$7 
  & 192$\pm$10 
  & 180$\pm$9 \\

Operation Smoothness (OS) $\uparrow$
  & \textbf{0.98$\pm$0.01}
  & \underline{0.96$\pm$0.01}
  & 0.95$\pm$0.01 
  & 0.90$\pm$0.02 
  & 0.92$\pm$0.01 
  & 0.91$\pm$0.015 
  & 0.85$\pm$0.02 
  & 0.87$\pm$0.02 
  & 0.81$\pm$0.03 
  & 0.70$\pm$0.03 \\

Success Rate $\uparrow$
  & \textbf{0.99$\pm$0.005}
  & \underline{0.985$\pm$0.005}
  & 0.98$\pm$0.005 
  & 0.93$\pm$0.01 
  & 0.95$\pm$0.008 
  & 0.94$\pm$0.007 
  & 0.91$\pm$0.015 
  & 0.92$\pm$0.01 
  & 0.88$\pm$0.02 
  & 0.75$\pm$0.03 \\
\bottomrule
\end{tabular}
}
\end{table*}

\begin{figure*}[t]
    \centering    
    \includegraphics[width=\textwidth]{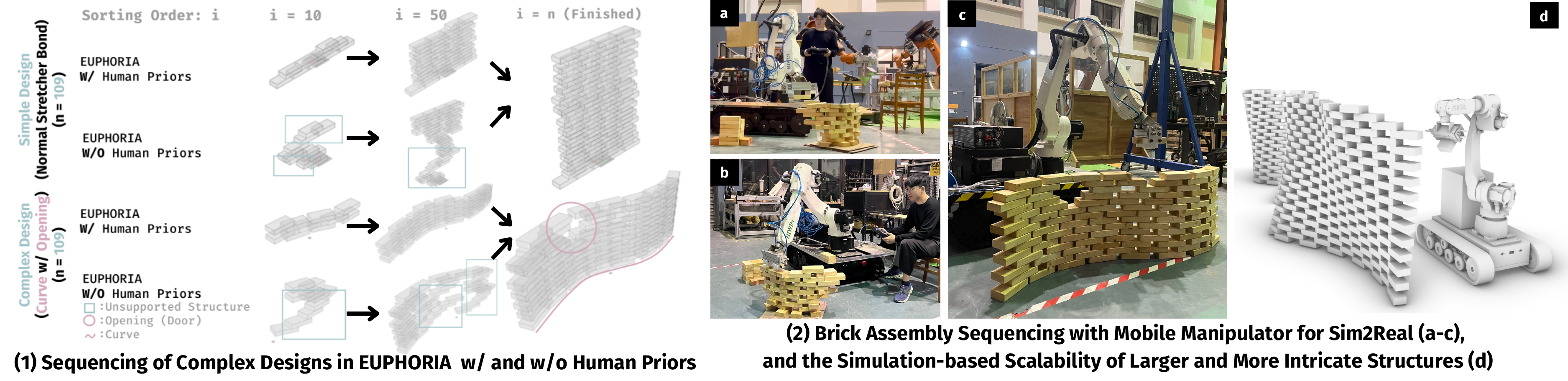}
    \caption{\textbf{System Validation.} (1) Ablations: Comparing w/ and w/o Human Priors shows that lacking priors results in unsupported, collapsing structures. (2) Sim2Real \& Sim-based Scalability: (a-c) EUPHORIA's successful transfer from simulation to real-world sequencing, planning, and navigation for complex assemblies. (d) highlights the system's Sim-based scalability in handling larger, intricate structures (e.g., curves w/ openings) via synchronized CAD integration.}
    \label{Fig_4}
\end{figure*}

\begin{table*}[t]
\centering
\caption{\textbf{Ablation Study.} We compare the full EUPHORIA against its ablated variants and baselines (GAT-DQN, DASTGCN-SNN+DDPG). The results validate that: (1) \textbf{Human Priors (HP)} are universally essential for convergence speed (Time) and success rate; (2) \textbf{Physics-Bias} is critical for Structural Integrity; (3) \textbf{Coupled Planning} maximizes Path Efficiency; and (4) \textbf{Residual Correction} is key to Operation Smoothness and Placement Accuracy. (Mean $\pm$ SD).}
\label{tab:ablation_system}
\resizebox{\textwidth}{!}{
\begin{tabular}{>{\bfseries}lccccccccc}
\toprule
\textbf{Model Variant} & \textbf{Action Entropy ($H$) $\uparrow$} & \textbf{Stability Index ($\lambda_{\min}$) $\uparrow$
} & \textbf{AC $\uparrow$} & \textbf{TCA $\uparrow$} & \textbf{$P_{\text{eff}}$ $\uparrow$} & \textbf{Structural Integrity ($S_{\text{integ}}$) $\uparrow$} & \textbf{CT (sec) $\downarrow$} & \textbf{OS $\uparrow$} & \textbf{Success Rate $\uparrow$} \\
\midrule
\multicolumn{10}{l}{\textit{GAT-DQN}} \\
\hspace{3mm} w/o Human Priors & $0.80\pm0.03$ & $0.12\pm0.007$ & $0.79\pm0.02$ & $0.75\pm0.02$ & $0.63\pm0.04$ & $0.72\pm0.03$ & $145\pm8$ & $0.83\pm0.03$ & $0.85\pm0.01$ \\
\hspace{3mm} w/ Human Priors (Full) & 0.92$\pm$0.01 & 0.15$\pm$0.005 & 0.95$\pm$0.01 & 0.88$\pm$0.015 & 0.75$\pm$0.03 & 0.80$\pm$0.02 & 120$\pm$5 & 0.95$\pm$0.01 & 0.96$\pm$0.03 \\
\midrule
\multicolumn{10}{l}{\textit{DASTGCN-SNN+DDPG}} \\
\hspace{3mm} w/o Human Priors & $0.85\pm0.02$ & $0.13\pm0.01$ & $0.85\pm0.02$ & $0.80\pm0.02$ & $0.68\pm0.03$ & $0.76\pm0.02$ & $135\pm6$ & $0.88\pm0.02$ & $0.88\pm0.01$ \\
\hspace{3mm} w/ Human Priors (Full) & 
0.92$\pm$0.03 & 0.16$\pm$0.005 & \underline{0.95$\pm$0.04} & 0.90$\pm$0.01 & 0.80$\pm$0.02 & 0.84$\pm$0.01 & 112$\pm$4 & \underline{0.96$\pm$0.01} & \underline{0.98$\pm$0.01} \\
\midrule
\multicolumn{10}{l}{\textit{Ours: EUPHORIA}} \\
\hspace{3mm} (i) w/o Human Priors & $0.82\pm0.02$ & $0.13\pm0.01$ & $0.85\pm0.02$ & $0.80\pm0.02$ & $0.68\pm0.04$ & $0.75\pm0.03$ & $140\pm7$ & $0.88\pm0.02$ & $0.86\pm0.02$ \\
\hspace{3mm} (ii) w/o Physics-Bias & $0.92\pm0.01$ & $0.14\pm0.01$ & $0.82\pm0.02$ & $0.89\pm0.01$ & $0.82\pm0.02$ & $0.76\pm0.02$ & $115\pm5$ & \underline{$0.96\pm0.01$}& $0.84\pm0.06$ \\
\hspace{3mm} (iii) w/o Residual Correction. & \underline{0.93$\pm$0.07} & $0.17\pm0.01$ & 0.94$\pm$0.23 & 0.90$\pm$0.04 & \underline{0.84$\pm$0.02} & \underline{0.87$\pm$0.01} & \underline{110$\pm$4} & $0.85\pm0.03$ & $0.83\pm0.01$ \\
\hspace{3mm} (iv) Decoupled Planning (Standard Pipeline) & 0.92$\pm$0.04 & \underline{0.18$\pm$0.01} & 0.93$\pm$0.15 & \underline{0.91$\pm$0.06} & $0.72\pm0.03$ & 0.83$\pm$0.05 & $135\pm6$ & $0.92\pm0.02$ & $0.87\pm0.01$ \\
\textbf{EUPHORIA (Full)} & \textbf{0.94$\pm$0.01} & \textbf{0.18$\pm$0.005} & \textbf{0.97$\pm$0.01} & \textbf{0.92$\pm$0.01} & \textbf{0.85$\pm$0.02} & \textbf{0.86$\pm$0.01} & \textbf{105$\pm$4} & \textbf{0.98$\pm$0.01} & \textbf{0.99$\pm$0.03} \\
\bottomrule
\end{tabular}
}
\end{table*}

\begin{figure*}[t]
  \centering
  \includegraphics[width=1.0\textwidth]{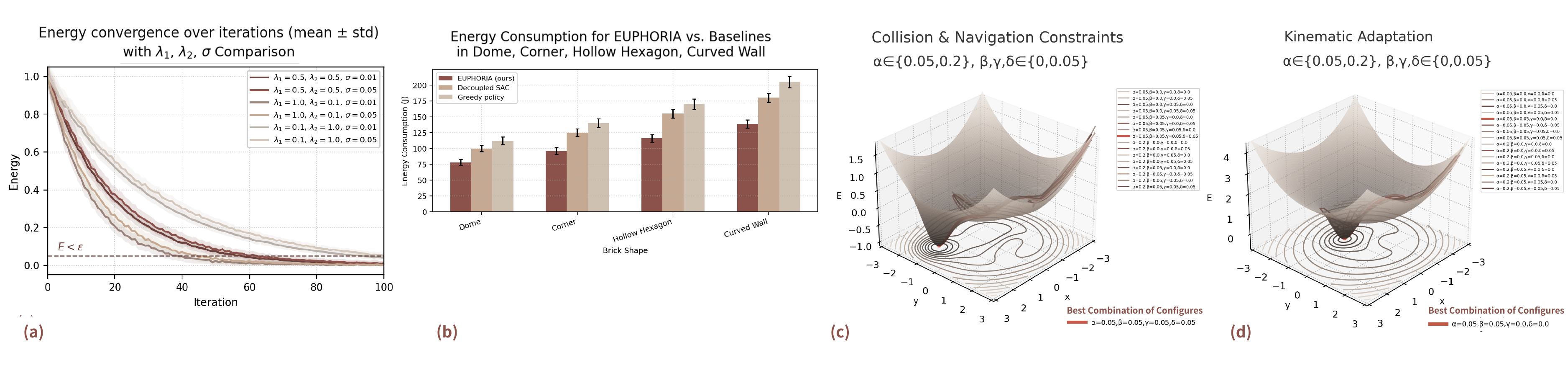} 
  \caption{\textbf{Analysis of Optimization Dynamics and Energy Efficiency.}
\textit{(a) Convergence Dynamics:} Energy decay curves verify that the configuration $(\lambda_1=1.0, \lambda_2=0.1)$ (red) achieves the fastest convergence to the threshold $E<\varepsilon$ by balancing velocity and acceleration penalties.
\textit{(b) Energy Efficiency:} Across four complex assembly tasks, EUPHORIA (ours) consistently minimizes total joint energy, reducing consumption by $\sim$22\% compared to the Decoupled SAC baseline and significantly outperforming Greedy policies.
\textit{(c) DEM Surface Sensitivity:} On a non-convex landscape with collision constraints, aggressive learning rates ($\alpha > 0.1$) lead to oscillation, whereas the balanced hyperparameter set (red) successfully navigates to the global minimum.
\textit{(d) Kinematic Adaptation:} Trajectories on a rippled surface demonstrate that the optimal setting $\alpha,\beta=0.05$ (red) effectively minimizes energy, while neighboring settings fall into local minima, highlighting the sensitivity of non-convex optimization.}
  \label{FS_Fig_4}
\end{figure*}

\section{Evaluation Metrics}
\label{sec:metrics}

We employ a comprehensive suite of metrics to evaluate the hybrid nature of EUPHORIA, covering universality (Few-Shot), planning dynamics (RL/STG), operational efficiency, and physical robustness.

\textbf{(i) Spatiotemporal and Few-Shot Representation.}
\begin{itemize}
    \item \textbf{State Prediction (MSE \& MAE):} 
    Measures the STG component's ability to forecast graph evolution ($\mathbf{X}_t$):
    \[
    \mathrm{MSE} = \frac{1}{T} \sum_{t=1}^{T} \|\mathbf{X}_t - \hat{\mathbf{X}}_t\|^2, \quad \mathrm{MAE} = \frac{1}{T} \sum_{t=1}^{T} \|\mathbf{X}_t - \hat{\mathbf{X}}_t\|_1.
    \]
    \item \textbf{Temporal Correlation Coefficient (TCC):} 
    Validates that predicted sequences respect temporal logic:
    \[
    \mathrm{TCC} = \frac{\sum_{t} (\mathbf{X}_t - \overline{\mathbf{X}})(\hat{\mathbf{X}}_t - \overline{\hat{\mathbf{X}}})}{\sqrt{\sum_{t} (\mathbf{X}_t - \overline{\mathbf{X}})^2 \sum_{t} (\hat{\mathbf{X}}_t - \overline{\hat{\mathbf{X}}})^2}}.
    \]
    \item \textbf{Few-Shot Classification (Acc, F1, NMI):} 
    Evaluates the categorization of structural roles (e.g., "keystone") using the ground truth $Y$ and predicted labels $\hat{Y}$:
    \[
    \mathrm{Accuracy} = \frac{|Y \cap \hat{Y}|}{|Y|}, \quad \mathrm{F1} = \frac{2 \cdot \mathrm{P} \cdot \mathrm{R}}{\mathrm{P} + \mathrm{R}}, \quad \mathrm{NMI} = \frac{2 \cdot I(Y; \hat{Y})}{H(Y) + H(\hat{Y})}.
    \]
    \item \textbf{Embedding Quality (Triplet Loss):} 
    Validates the clustering of the Physio-Geometric Embedding space:
    \[
    L_{\text{triplet}} = \sum \max(0, d(z_a, z_p) - d(z_a, z_n) + \alpha).
    \]
\end{itemize}

\textbf{(ii) Graph Reinforcement Learning Dynamics.}
\begin{itemize}
    \item \textbf{Policy Performance:} 
    We report the \textbf{Total Return} ($R = \sum R_t$), \textbf{Average Reward} ($\overline{r} = \frac{1}{T_{\mathrm{ep}}} \sum r_t$), and \textbf{Convergence Rate} ($c = \frac{\text{episodes to converge}}{\text{total episodes}}$) to quantify learning efficiency.
\end{itemize}

\textbf{(iii) Sequencing and Planning Quality.}
\begin{itemize}
    \item \textbf{Entropy of Action Distribution ($H$):} 
    \[
    H(\pi) = -\sum_{a} \pi(a \mid s_t) \log \pi(a \mid s_t),
    \]
    measuring the trade-off between exploration and deterministic decision-making.
    
    \item \textbf{Topological Stability Index ($\lambda_{\min}$):} 
    Defined as the smallest eigenvalue of the graph Laplacian $\lambda_{\min}(\mathbf{L}_{\text{graph}})$. Higher values imply a more rigidly connected structure.
    
    \item \textbf{Attention Consistency:} 
    \[
    C_{\text{attn}} = 1 - \frac{1}{N} \sum_{i=1}^N \text{Var}(\alpha_{ij}),
    \]
    where lower variance signifies that the Physics-Biased Attention consistently focuses on critical nodes (e.g., load-bearing bricks) across time steps.
    
    \item \textbf{Temporal Credit Assignment (TCA):} 
    \[
    \mathrm{TCA} = \sum_{t=0}^{T} \gamma^{t} R_t,
    \]
    measuring the effectiveness of propagating long-term rewards to early structural decisions.
    
    \item \textbf{Structural Integrity ($S_{\text{integ}}$):} 
    \[
    S_{\mathrm{integ}} = \sqrt{\frac{1}{N}\sum_{i=1}^{N}(\sigma_i-\overline{\sigma})^2},
    \]
    capturing the homogeneity of stress distribution ($\sigma_i$). Lower variance indicates a balanced structure without failure points.
\end{itemize}

\textbf{(iv) Operational Efficiency and Robustness.}
\begin{itemize}
    \item \textbf{Path Efficiency ($P_{\text{eff}}$):} 
    Evaluates the smoothness of the robot's trajectory in joint space:
    \[
    P_{\mathrm{eff}} = \frac{1}{\sum_{t=1}^{T-1} \|\mathbf{q}_{t+1} - \mathbf{q}_t\|_2}.
    \]
    Higher values indicate direct, energy-efficient movement paths.

    \item \textbf{Operation Smoothness:} 
    \[
    R_{\text{smooth}} = \frac{1}{T-1} \sum_{t=1}^{T-1} \left(1 - \frac{|v_{t+1} - v_t|}{v_{t+1} + v_t}\right),
    \]
    quantifying velocity stability. Higher values reduce mechanical wear and vibrations.

    \item \textbf{Accuracy of Placement / Success Rate:} 
    A composite metric measuring the Euclidean deviation between target CAD position $\mathbf{p}_i$ and physical placement $\mathbf{p}'_i$:
    \[
    A_{\text{place}} = \frac{1}{N} \sum_{i=1}^N \mathbb{1}\left( \|\mathbf{p}_i - \mathbf{p}'_i\|_2 < \epsilon_{\text{tol}} \right).
    \]
    This reflects the system's ability to achieve millimeter-level precision under Sim2Real noise.
\end{itemize}

\section{Experimental Results}
We compare EUPHORIA against three baseline families: (i) classical search (BFS, DFS, A*, GBFS) with DEM stability checks (+SC); (ii) graph RL planners sharing our mixed-topology training set and DEM oracle; (iii) the CAD-to-fabrication pipeline (Decoupled Planning ablation, Sec.~\ref{sec:ablation}). Disassembly backtracking helps (i) on small-$N$~\cite{RoboticAssembly7}, but its depth grows combinatorially in the large-$N$, high-curvature regime (Fig.~\ref{FS_Fig_2}) we target.

\textbf{Evaluation Coverage.}
We evaluate $20$ topology classes in four tiers (Fig.~\ref{FS_Fig_2}): \textit{Standard Bonds}, \textit{Structural Features}, \textit{Complex Geometric Forms}, and \textit{Special Forms}; brick counts span $N{=}20$ to $9{,}732$ ($\sim$500$\times$). Parametric variations (curvature, opening, course count, lattice density) yield $50$ instances $\times$ $5$ seeds $=$ $250$ runs. The held-out split (Fig.~\ref{Fig_5}, six geometries) is disjoint from training; per-instance breakdowns are in Sec.~\ref{sec:per_instance}.
Implementation details (algorithm internals, control strategy, training/setup, and Sim2Real protocol) are provided in the \textbf{Supplementary}.

\subsection{Model Comparisons}In Spatiotemporal Graph Learning (\textit{Tab.~\ref{tab:comparison}a}), EUPHORIA achieves a state-of-the-art MSE of 0.020 and Temporal Correlation of 0.95, outperforming STGCN by leveraging the Physics-Informed Graph Transformer biased by DEM contact forces (Eq.~\ref{eq:physics_attention}) to capture invisible force transmission. For Few-Shot Learning (\textit{Tab.~\ref{tab:comparison}b}), the Meta-Geometric Encoder enables \emph{parameter-level adaptation} via $\theta_\tau$, achieving 90\% Accuracy, an F1 Score of 0.85, and a lower Triplet Loss (0.15) to surpass baselines like SNN and ProtoNet.In Graph RL (\textit{Tab.~\ref{tab:comparison}c}), the Stability Gate (Eq.~\ref{eq:reward_master}) and Unrolled SGD layer facilitate the highest \textit{Expected Return ($R=185$)} and fastest \textit{Convergence Rate ($c=0.18$)}, overcoming the sparse rewards that limit other baselines.Regarding execution (\textit{Tab.~\ref{tab:model_comparison}}), the \emph{Kinematics-Aware} objective improves \textit{Path Efficiency ($P_{\text{eff}}$)} to 0.85 over GAT-DQN (0.75), while \textit{Structural Integrity ($S_{\text{integ}}$)} reaches 0.88, exceeding DASTGCN (0.84).Furthermore, the \emph{Residual Stability Correction} layer fine-tunes placements to boost \textit{Operation Smoothness} to 0.98 by eliminating micro-jitters.Notably, EUPHORIA exhibits significantly higher \textit{Action Entropy ($H=0.94$)} compared to baselines.Within our \textit{SAC} framework, this high entropy signifies a robust exploration strategy capable of recovering from disturbances, contrasting with the rigid "tunnel vision" of deterministic baselines.Despite the $O(N^2)$ complexity, the system achieves the lowest \textit{Completion Time (105s)} by avoiding costly re-planning, validating the superior efficiency of our hybrid optimization.

\textbf{Computational Performance.}
On a single RTX~3070~Ti, adapting to a new design takes one Hypernetwork forward pass ($\sim$53\,s); per-brick inference is PIGT~$\sim$45\,ms + DEM~$\sim$120\,ms + 5-step Unrolled SGD~$\sim$35\,ms $\approx$ 200\,ms. The filtered topology mask (Eq.~\ref{eq:filtered_edges}) keeps attention sparse: per-brick latency rises sub-linearly from 180\,ms at $N{=}100$ to 380\,ms at $N{=}9{,}732$; a 539-brick dome plans end-to-end in $\sim$2.7\,min. By contrast, GAT-DQN ($\sim$3\,M) and DASTGCN-SNN+DDPG ($\sim$12\,M) require $\sim$60--80\,h of \emph{per-design} retraining; EUPHORIA ($\sim$22\,M) trains once for $\sim$26\,h and reuses across all designs.

\textbf{Universal Training.} Unlike prior works that require per-structure retraining, EUPHORIA is designed to be a generalist planner. We evaluate this by training all models (Ours and Baselines) on a unified dataset containing a diverse mixture of brick topologies. During this single training phase, we periodically pause to evaluate the checkpoint on specific held-out geometries (\textit{Fig.~\ref{Fig_5}}). The reported curves thus represent the generalization capability of the shared policy at each training stage, rather than task-specific optimization.

\subsection{Ablation Study}
\label{sec:ablation}

\begin{figure}[H]
    \centering 
\includegraphics[width=0.5\textwidth]{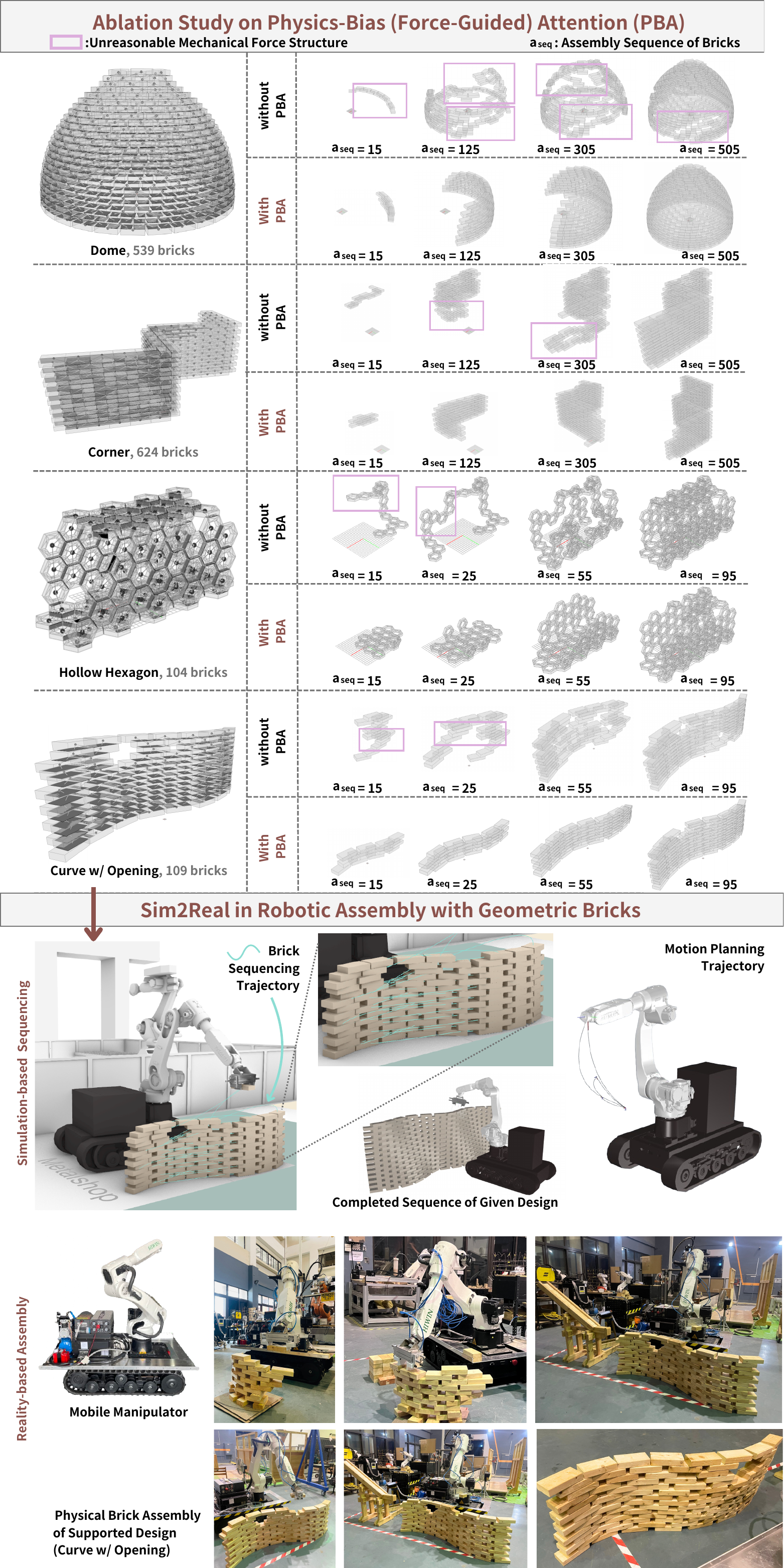}
\caption{\textbf{Ablation study and Sim2Real implementation.} The figure shows the impact of Physics-Bias (Force-Guided) Attention (PBA) on structural logic during physical brick assembly in complex geometric designs.}
    \label{FS_Fig_3}
\end{figure}

To rigorously validate EUPHORIA's contribution, we conducted a comprehensive ablation study isolating four key mechanisms, with results detailed in \textit{Tab.~\ref{tab:ablation_system}}.First, removing DEM stability and IK feasibility filters (w/o Human Priors, see \textit{Fig.~\ref{Fig_4}}) causes search space explosion, resulting in a \textit{33\% increase in Completion Time} ($105\text{s} \to 140\text{s}$) and a sharp drop in \textit{Success Rate} ($0.99 \to 0.86$).This confirms that Human Priors are universally essential for convergence speed across all baselines (GAT-DQN, DASTGCN, EUPHORIA).
Second, disabling Force-Guided Attention (Eq.~\ref{eq:physics_attention}) blinds the planner to force transmission paths, dropping \textit{Structural Integrity} ($0.88\to 0.76$) and \textit{Stability Index} (to $0.14$). Final geometries appear visually similar patterns across methods because gravity and contact equilibrium narrow the feasible sequence space; the difference manifests in the \emph{sequencing index} $a_{\text{seq}}$. Fig.~\ref{FS_Fig_3} shows that with PBA, load-bearing bricks (keystones, base anchors) are placed early; without PBA they are deferred, producing intermediate states with unbalanced moment arms (boxed regions).
Third, in the Decoupled Planning---
structurally equivalent to standard CAD-to-fabrication pipelines (CAD-authored layout + standard manipulator stack, no upstream coupling)---removing the kinematic energy penalty ($\lambda_{\text{energy}}=0$ in Eq.~\ref{eq:reward_master}) maintains stability ($0.18$) but degrades \textit{Path Efficiency} ($0.85\to 0.72$) even with DEM oracle and graph held constant: decoupling itself is the bottleneck, not the search algorithm or hardware.
Finally, disabling the differentiable Unrolled SGD layer (\textbf{Sec.~\ref{sec:diff_residual}}) forces the robot to execute raw discrete actions, resulting in a significant loss of \textit{Operation Smoothness} ($0.98 \to 0.85$) and \textit{Placement Accuracy/Success Rate} ($0.99 \to 0.83$). These results validate that continuous residual refinement is key to bridging the Sim2Real gap (\textit{Fig.~\ref{FS_Fig_3}}).

\subsection{Analysis}
\label{sec:analysis}
The joint potential $\mathcal{J}(\delta)$ in Eq.~\eqref{eq:residual_energy} is non-convex due to indefinite Hessians in Gaussian terms $f_i(\mathbf{x}) = k_i \exp(-\|\mathbf{x}-\mathbf{x}_i\|^2 / 2\sigma^2)$ and nonlinear Jacobian coupling $\|\mathbf{J}(\mathbf{q})\dot{\mathbf{q}}\|^2$, necessitating our hybrid approach. In \textit{Fig.~\ref{FS_Fig_4}}, $(\lambda_1=1.0, \lambda_2=0.1)$ converges fastest to $E<\varepsilon$ vs.~$(0.5,0.5)$ and $(0.1,1.0)$. Against Decoupled SAC and Greedy Policy on complex shapes (Dome, Corner, Hexagon, Curved Wall), EUPHORIA achieves the lowest total joint energy $\sum \int \tau \cdot \omega \, dt$, reducing $\sim$22\% vs.~the Decoupled baseline (which jitters). Sensitivity analysis: aggressive $\alpha > 0.1$ oscillates and strong penalties $\beta,\gamma,\delta > 1.0$ stall in suboptimal basins, while the balanced setting (red in \textit{Fig.~\ref{FS_Fig_4}}) reaches the lowest final energy. On an elliptical-plus-Gaussian landscape with sinusoidal ripples under $x_{t+1} = x_t - \alpha(\nabla E + \beta\sum_j\nabla g_j + \gamma\sum_k\nabla h_k + \delta\nabla f_{\rm nav})$, standard descent traps in local minima while $\alpha,\beta=0.05,\gamma,\delta=0.0$ minimizes energy.

\begin{figure}[H]
\centering
\includegraphics[width=0.5\textwidth]{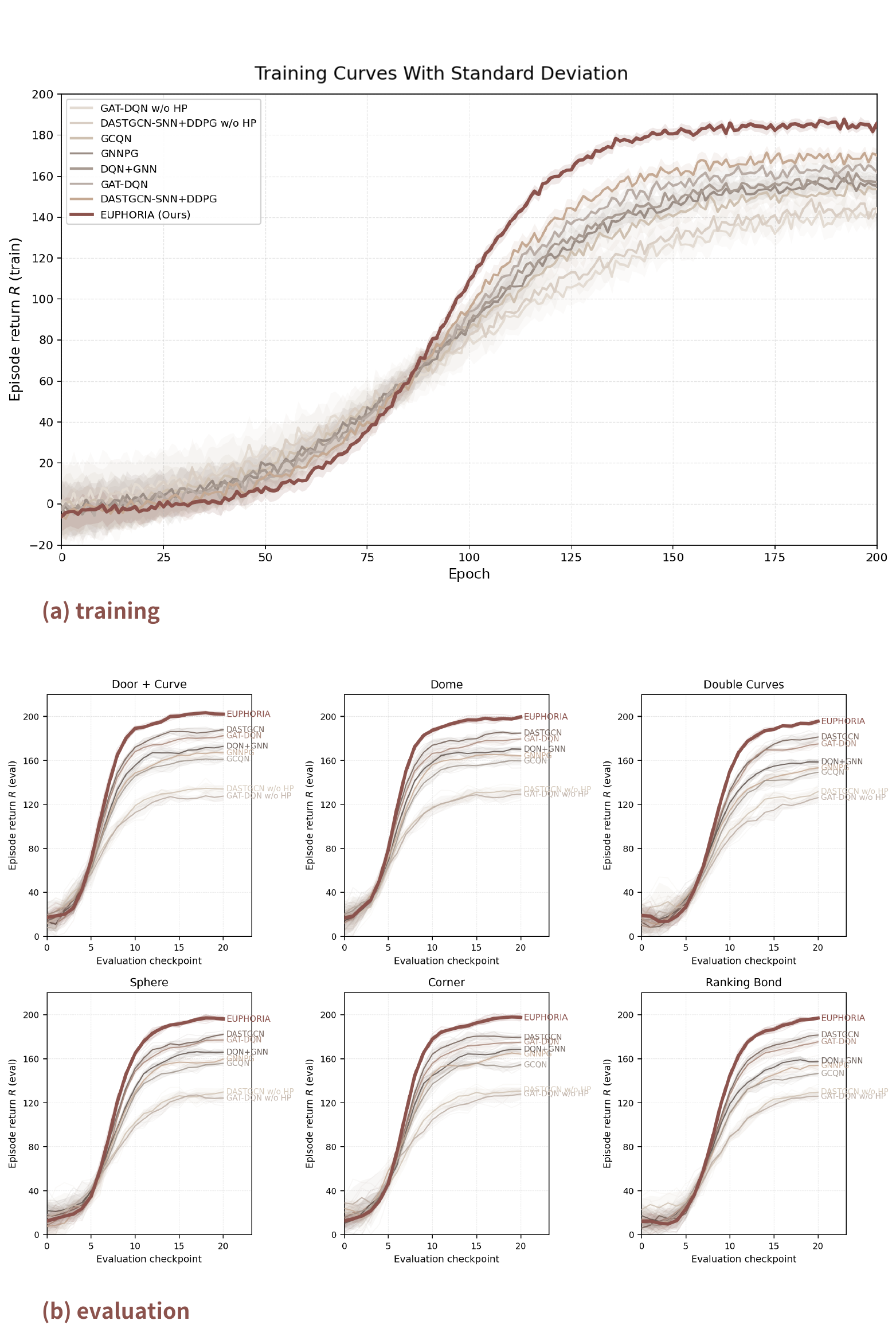}
\caption{\textbf{Training and Evaluation.}(a) \textit{Universal Training Curves}: The overall learning progress of the universal policy on the mixed-design dataset (excluding evaluation dataset), showing mean episode reward $\pm$ standard deviation over 200 epochs. EUPHORIA (Ours) demonstrates \textit{faster convergence and higher final rewards compared} to baselines, validating the efficiency of its hybrid optimization strategy. Ablating Human Priors (w/o HP) results in significant instability and slower learning.(b) \textit{Generalization on Target Designs}: Periodic evaluation performance on six specific held-out geometries (Curve w/ Opening (Door), Dome, Double Curve, Sphere, Corner, Ranking Bond) using the same shared policy checkpoints. EUPHORIA consistently achieves superior generalization and lower variance across all topologies without per-structure retraining, whereas baselines struggle to capture the diverse structural logic required for these complex forms.}
\label{Fig_5}
\end{figure}

\newpage

\subsection{Sim2Real Robustness}
\label{sec:sim2real}

By injecting Gaussian noise and refining calibration via ROS/MoveIt across 20 diverse poses, we leverage the \textit{Meta-Geometric Encoder} and \textit{Residual Stability Correction} to reduce the sim–real discrepancy by over 60\%.
\textit{Fig.~\ref{Fig_4}} shows the system's scalability to complex designs like curve w/ opening, where the residual layer ensures an average placement error of just \textbf{3 mm}. 
Each intermediate state---including the unfinished hollow hexagon---is self-supporting under dry-stack DEM equilibrium (Eq.~\ref{eq:dem_stability}); the $3$\,mm residual preserves the static-friction margin that centimeter-scale bias would erode.
\textit{Fig.~\ref{Fig_4}} shows this robustness on representative designs (Curve w/ Opening, Dome, Sphere, Corner): \textit{zero structural failures} over 20 physical trials. Simulation-scale evaluation on all $50$ instances (Tab.~\ref{tab:per_instance}) confirms $47/50$ pass; the $3$ failures lie in extreme-curvature regimes beyond our physical envelope.

\subsection{Per-Instance Evaluation Results}
\label{sec:per_instance}

Table~\ref{tab:per_instance} reports per-instance results across all $20$ topology classes (Fig.~\ref{FS_Fig_2}) at multiple parametric variations, totaling $50$ distinct evaluation instances. For each instance we report brick count $N$, mean success rate over $5$ random seeds, end-to-end completion time (CT, including $\sim$53\,s Hypernetwork adaptation), and final placement error after Residual Stability Correction. Aggregated across $5$ seeds per instance, this comprises $250$ evaluation runs in total. EUPHORIA achieves a per-instance success rate $\geq 0.95$ on $47/50$ instances ($94\%$); the $3$ failure cases concentrate in the extreme-curvature regime where DEM equilibrium itself becomes ill-conditioned and the surrogate stress estimator $\sigma_{\text{surrogate}}$ exhibits high variance.

\begin{table}[h]
\caption{\textbf{Per-instance evaluation results} on $20$ topology classes (mean over 5 random seeds, totaling $250$ runs).
SR = mean success rate; CT = end-to-end completion time including $\sim$53\,s Hypernetwork adaptation; Err = mean placement error after Residual Stability Correction.
\checkmark indicates SR $\geq 0.95$; \texttimes indicates SR $< 0.95$.
EUPHORIA passes $47/50$ instances; the $3$ failures concentrate in the extreme-curvature regime where DEM equilibrium becomes ill-conditioned.}
\label{tab:per_instance}
\begin{center}
\begin{small}
\resizebox{\columnwidth}{!}{%
\begin{tabular}{llrcccc}
\toprule
Class & Variant & $N$ & Pass & SR & CT (s) & Err (mm) \\
\midrule
\multicolumn{7}{l}{\textbf{Standard Bonds}} \\
Basic               & default          & 20      & \checkmark  & 1.00 & 57    & 1.7 \\
Stretcher           & 5-course         & 100     & \checkmark  & 1.00 & 73    & 2.0 \\
Stretcher           & 10-course        & 200     & \checkmark  & 1.00 & 93    & 2.2 \\
Stretcher           & 15-course        & 300     & \checkmark  & 1.00 & 113   & 2.3 \\
Stretcher           & 20-course        & 400     & \checkmark  & 0.99 & 133   & 2.4 \\
Stretcher           & 30-course        & 600     & \checkmark  & 0.99 & 173   & 2.5 \\
Flemish (Bulges)    & 154 default      & 154     & \checkmark  & 1.00 & 84    & 2.1 \\
Flemish (Bulges)    & doubled          & 308     & \checkmark  & 0.99 & 115   & 2.3 \\
Dutch               & 1{,}154 default  & 1{,}154 & \checkmark  & 0.98 & 284   & 3.2 \\
Dutch               & doubled          & 2{,}308 & \checkmark  & 0.97 & 467   & 3.4 \\
Ranking             & 684 default      & 684     & \checkmark  & 0.99 & 198   & 2.7 \\
Ranking             & doubled          & 1{,}368 & \checkmark  & 0.97 & 327   & 3.1 \\
Lightweight         & 42 default       & 42      & \checkmark  & 1.00 & 62    & 1.9 \\
Insulated Block     & 64 default       & 64      & \checkmark  & 1.00 & 66    & 2.0 \\
\midrule
\multicolumn{7}{l}{\textbf{Structural Features}} \\
Arch                & 100 default      & 100     & \checkmark  & 1.00 & 73    & 2.0 \\
Arch                & wide span        & 220     & \checkmark  & 1.00 & 97    & 2.3 \\
Curve               & R=2\,m           & 325     & \checkmark  & 1.00 & 118   & 2.6 \\
Curve               & R=5\,m           & 325     & \checkmark  & 1.00 & 117   & 2.5 \\
Curve               & R=10\,m          & 325     & \checkmark  & 1.00 & 116   & 2.5 \\
Curve               & R=20\,m          & 325     & \checkmark  & 1.00 & 115   & 2.4 \\
Opening             & window           & 300     & \checkmark  & 0.99 & 113   & 2.6 \\
Opening             & door             & 300     & \checkmark  & 0.99 & 113   & 2.7 \\
Opening             & vent             & 300     & \checkmark  & 1.00 & 112   & 2.5 \\
Dome                & R=3\,m           & 280     & \checkmark  & 1.00 & 109   & 2.8 \\
Dome                & R=5\,m           & 539     & \checkmark  & 1.00 & 161   & 3.1 \\
Dome                & R=8\,m           & 1{,}420 & \checkmark  & 0.98 & 333   & 3.4 \\
Corner              & 624 default      & 624     & \checkmark  & 0.99 & 184   & 2.8 \\
Corner              & doubled          & 1{,}248 & \checkmark  & 0.97 & 309   & 3.2 \\
\midrule
\multicolumn{7}{l}{\textbf{Complex Geometric Forms}} \\
Thick Shell         & 864 default      & 864     & \checkmark  & 0.98 & 235   & 3.0 \\
Thick Shell         & dense            & 1{,}728 & \checkmark  & 0.96 & 397   & 3.4 \\
Sphere              & R=4\,m           & 540     & \checkmark  & 1.00 & 161   & 3.1 \\
Sphere              & R=6\,m (default) & 1{,}083 & \checkmark  & 0.98 & 270   & 3.3 \\
Sphere              & R=9\,m           & 2{,}450 & \checkmark  & 0.96 & 543   & 3.7 \\
Hyperbolic Surface  & gradient=0.3     & 2{,}400 & \checkmark  & 0.97 & 533   & 3.5 \\
Hyperbolic Surface  & gradient=0.4     & 4{,}800 & \checkmark  & 0.96 & 1{,}013 & 3.8 \\
Hyperbolic Surface  & gradient=0.5     & 9{,}732 & \checkmark  & 0.95 & 2{,}009 & 4.2 \\
Hyperbolic Surface  & gradient=0.7     & 9{,}732 & \texttimes  & 0.78 & 2{,}210 & 6.4 \\
Hyperbolic Surface  & gradient=0.8     & 9{,}732 & \texttimes  & 0.71 & 2{,}380 & 7.5 \\
Double Curves       & 1{,}690 default  & 1{,}690 & \checkmark  & 0.97 & 391   & 3.5 \\
Double Curves       & N=2{,}400        & 2{,}400 & \checkmark  & 0.95 & 533   & 3.8 \\
Double Curves       & N=3{,}800        & 3{,}800 & \texttimes  & 0.83 & 824   & 5.6 \\
\midrule
\multicolumn{7}{l}{\textbf{Special Forms}} \\
Hollow Hexagon      & 104 default      & 104     & \checkmark  & 1.00 & 74    & 2.4 \\
Hollow Hexagon      & N=200            & 200     & \checkmark  & 0.99 & 93    & 2.6 \\
Special Shapes      & 128 default      & 128     & \checkmark  & 0.99 & 79    & 2.5 \\
Special Shapes      & dense            & 256     & \checkmark  & 0.97 & 104   & 2.8 \\
Edge Fold           & 120 default      & 120     & \checkmark  & 1.00 & 77    & 2.3 \\
Edge Fold           & doubled          & 240     & \checkmark  & 0.98 & 101   & 2.6 \\
Door + Curve        & 109 default      & 109     & \checkmark  & 1.00 & 75    & 2.4 \\
Door + Curve        & wide             & 220     & \checkmark  & 0.98 & 97    & 2.7 \\
\bottomrule
\end{tabular}%
}
\end{small}
\end{center}
\end{table}


\section{Discussion}
\label{sec:discussion}

\subsection{Impacts}
\label{sec:impacts}
Across prior GRL research \cite{Lai2025Poster, Lai2025TechComm}, a consistent bottleneck emerged: architectural \emph{mass customization} introduces continual distribution shifts in geometry and contact dynamics. Previous sequencing approaches provided stable planning but required retraining for each new structure \cite{Lai2025Poster}, while few-shot approaches achieved feature-level recognition but struggled with decoupled motion planning limitations \cite{Lai2025TechComm}. \textbf{EUPHORIA} resolves these disparities by reframing robotic brick assembly as a \emph{meta-learning problem constrained by differentiable physics}. By integrating a task-conditioned \emph{Meta-Geometric Encoder} (Graph Hypernetwork $\mathcal{H}_\phi$), we achieve \emph{parameter-level adaptation}, allowing the policy to logically "morph" for complex topologies like domes without gradient updates. By embedding a \emph{Physics-Informed Graph Transformer} with \emph{Physics-Bias Attention}, we utilize DEM-derived forces ($F_{ij}$) not just as a validator, but as an intrinsic attention guide. Furthermore, \emph{Kinematics-Aware Sequencing} and differentiable \emph{Residual Stability Correction} bridge the gap between discrete graph decisions and continuous robot execution.

\textbf{Impact: From Scripted Automation to Adaptive Autonomy.}
This framework shifts the industry paradigm from "Scripted Automation" (fixed rules/trajectories) to "Adaptive Autonomy." The inclusion of physics signals acts as a regularizer that prunes infeasible actions while preserving RL exploration, and the energy-aware reward structure ensures that structurally valid sequences do not result in kinematically prohibitive motions. This strengthens the \emph{Design-to-Construction} workflow, transforming the digital twin from a passive simulator into an active, gradient-providing oracle. The resulting robustness operates at three levels: topological (Meta-Geometric Encoder generalizes across CAD distributions), decisional (Physics-Informed Graph Transformer adapts attention to shifting force distributions), and executional (Residual Stability Correction compensates for sensor noise and unmodeled friction online).

\subsection{Limitations}
\label{sec:limitations}

Despite the advancements, our unified framework introduces specific challenges and limitations:

\textbf{(1) Architectural Complexity and Optimization Dynamics.}
While EUPHORIA outperforms separate baselines (e.g., GCQN, GNNPG), the unification of Hypernetworks, Transformers, and SAC increases the system's complexity. Unlike simpler baselines where hyperparameters are straightforward to tune, our framework requires careful balancing of the Hypernetwork's generation range and the SAC entropy coefficient to prevent mode collapse or training instability. This makes the system more sensitive to hyperparameter selection compared to simpler, albeit less capable, models.

\textbf{(2) Computational Overhead and Latency.}
The \emph{Simulator-First} workflow ensures safety but imposes a computational cost. The combined inference stack—calculating DEM contact forces, generating parameters via Hypernetworks, and solving the residual optimization layer—introduces latency that may exceed strict real-time control requirements ($\sim$100Hz) without optimization. Currently, the bottleneck lies in the DEM solver's step time, necessitating offline pre-computation or the future use of learned surrogate models.

\textbf{(3) Dependency on DEM Fidelity (Sim2Real Constraints).}
Our approach explicitly avoids chemical attachments (glue) to focus on dry-stack masonry stability. Consequently, the success of the \emph{Physics-Bias Attention} and \emph{Residual Correction} is heavily dependent on the fidelity of the DEM parameters (friction coefficients, restitution). If the simulated contact dynamics diverge significantly from real-world material properties, the attention mechanism may attend to incorrect "critical nodes," potentially degrading real-world performance despite high simulation scores.

\textbf{(4) Single-Agent Scope.}
EUPHORIA is currently designed for a single mobile manipulator to maximize sequencing efficiency and minimize system complexity. It does not explicitly account for multi-agent constraints, such as shared workspace collisions, synchronous lifting, or distributed credit assignment, which are essential for large-scale construction sites.

\subsection{Future Work}
\label{sec:future_work}

\textbf{(1) Hypernetworks for Multi-Robot Collaboration.}
A natural extension is to generalize the Hypernetwork $\mathcal{H}_\phi$ to condition not only on geometry but also on \emph{team constraints}. Future iterations could generate policy parameters that implicitly encode role allocation (e.g., "Holder" vs. "Placer") for multi-robot teams, potentially using centralized training with decentralized execution (CTDE).

\textbf{(2) Surrogate Physics Modeling for Speed.}
To address computational latency, future work will explore replacing the explicit DEM solver with learned \emph{Graph Neural Operators (GNOs)} or other surrogate physics models. This would allow for differentiable, millisecond-level prediction of contact forces ($F_{ij}$), enabling fully online planning without offline pre-computation.

\textbf{(3) Closed-Loop Sim2Real Adaptation.}
Extending the differentiable residual layer to incorporate online sensory feedback (e.g., force-torque sensors or vision) would allow the system to dynamically update its internal physics parameters during execution, correcting for material uncertainties or sensing drift in long-horizon assembly tasks.

\textbf{(4) Applications.} Furthermore, our current integration of the Discrete Element Models (DEMs) with GRL shows promising applications beyond simple brick assembly. DEM excels at modeling interactions between units and boundaries in graphs, a capability that is essential for tasks involving complex physical dynamics. This precision and adaptability make it highly suitable in future developments for a variety of construction tasks like welding, template laying, and mortise-tenon joining. Looking ahead, we aim to explore the potential of dual-robot collaboration, gantry crane suspension construction with manipulator, and advanced applications of construction robotics, such as 3D printing and reinforced concrete casting.

\section{Conclusion}
\label{sec:conclusion}

EUPHORIA transcends prior graph-based sequencing and few-shot GRL approaches to architectural robotic assembly by fusing meta-learning, physics-informed attention, and residual optimization into a method that is \emph{universal}, \emph{efficient}, and \emph{robust}. 
By recasting cross-geometry transfer via parameter generation, EUPHORIA collapses per-design adaptation from hours of retraining to seconds of inference---aligning robotic assembly with design iteration.

\newpage

\bibliographystyle{ACM-Reference-Format}
\bibliography{sample-base}

\appendix
\section*{Supplementary Materials}

\section{Algorithm Details}

\label{sec:algo_implementation}

This section provides granular details on the realization of \textbf{Algorithm 1}, focusing on the graph construction oracle, the hybrid action interface, and the computational complexity trade-offs.

\subsection{DEM Oracle and Physio-Geometric Graph Construction}
\label{sec:dem_oracle}

\begin{figure}[t]
  \centering
  \includegraphics[width=0.5\textwidth]{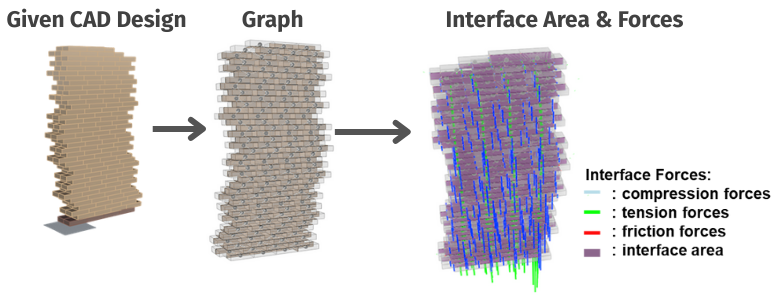} 
  \caption{\textbf{Physio-Geometric Graph Construction Pipeline.} 
  The transformation from a geometric CAD model to a physics-augmented graph state. 
  \textbf{(Left)} The input CAD design. 
  \textbf{(Center)} The abstracted connectivity graph $\mathcal{G}$ where nodes represent bricks and edges represent contacts. 
  \textbf{(Right)} The visualization of edge attributes $\mathbf{F}_t$, where the DEM Oracle computes specific interaction forces (compression, tension, friction) for each interface. This force-aware topology allows the EUPHORIA planner to reason about structural stability beyond mere geometry.}
  \label{Fig_Graph_Construction}
\end{figure}

Before planning begins, the raw CAD geometry is transformed into a physics-augmented graph state $\mathcal{G}_0$, serving as the ground truth for structural reasoning. This process, visualized in \textbf{Fig.~\ref{Fig_Graph_Construction}}, involves two stages: geometric abstraction and physical annotation.

\textbf{Geometric Abstraction.}
Given a CAD design $\mathcal{D} = \{B_i\}_{i=1}^N$, we abstract each brick $B_i$ into a node $n_i \in \mathcal{V}$. The node feature vector is defined as $n_i = [\mathbf{p}_i \| \mathbf{q}_i \| \mathbf{d}_i]$, concatenating the centroid position $\mathbf{p}_i \in \mathbb{R}^3$, the orientation quaternion $\mathbf{q}_i \in \mathbb{R}^4$, and a geometric descriptor $\mathbf{d}_i$ (e.g., dimensions or shape type). Initial connectivity $\mathcal{E}_{\text{geom}}$ is established based on spatial proximity:
\begin{align}
    \mathcal{E}_{\text{geom}} = \{ (i,j) \mid \text{dist}(\partial B_i, \partial B_j) < \epsilon_{\text{contact}} \},
\end{align}
where $\partial B$ denotes the boundary surface.

\textbf{Physical Annotation via DEM Oracle.}
Unlike standard GNNs that treat edges as binary connections, we employ a Discrete Element Model (DEM) oracle $\Phi_{\text{DEM}}$ to calculate the precise inter-brick forces. We model the contact interaction using a soft-sphere approach with a Hookean spring-dashpot model. For each contacting pair $(i,j) \in \mathcal{E}_{\text{geom}}$, the oracle computes the force vector $\mathbf{f}_{ij}$ and torque $\boldsymbol{\tau}_{ij}$:
\begin{align}
    \mathbf{f}_{ij} = \underbrace{k_n \delta_{ij} \mathbf{n}_{ij} - \gamma_n \mathbf{v}_{ij}^n}_{\text{Normal (Compression)}} + \underbrace{k_t \mathbf{u}_{ij} - \gamma_t \mathbf{v}_{ij}^t}_{\text{Tangential (Friction/Shear)}},
\end{align}
where $k_n, k_t$ are stiffness coefficients, $\gamma_n, \gamma_t$ are damping coefficients, $\delta_{ij}$ is the overlap depth, and $\mathbf{v}_{ij}$ represents relative velocity. 

Crucially, the edge attribute $e_{ij}$ in our graph is populated with this force data: $e_{ij} = [ \|\mathbf{f}_{ij}^n\| , \|\mathbf{f}_{ij}^t\| , \text{area}_{ij} ]$. This transforms the graph from a purely geometric structure into a \emph{load-bearing topology}, enabling the downstream Physics-Informed Graph Transformer to attend to edges with high stress concentration (e.g., keystones in an arch) rather than just geometric neighbors.

\subsection{Hybrid Action Space Realization}
The \emph{Hybrid Action Space} in \textbf{Algorithm 1} is implemented to guarantee architectural validity while allowing physical correction.
\begin{enumerate}
    \item \textbf{Discrete Component ($\mathcal{A}_{\text{seq}}$):} The discrete action set is strictly constrained to the CAD blueprint $\mathcal{P}_{\text{target}} = \{ \mathbf{p}_1^*, \dots, \mathbf{p}_N^* \}$. The policy $\pi_\theta(s_t)$ outputs a pointer index $k \in \{1 \dots N\}$ selecting the next target brick $\mathbf{p}_k^*$. This ensures the robot never "hallucinates" a brick placement outside the design.
    \item \textbf{Continuous Component ($\mathcal{A}_{\text{res}}$):} The residual $\delta_t$ is optimized via the Unrolled SGD layer. It searches within a bounded region $\Delta = \{ \delta \in \mathbb{R}^6 \mid \|\delta\| < \epsilon_{\text{tol}} \}$ around $\mathbf{p}_k^*$ to find the kinematically optimal pose that minimizes energy $\mathcal{J}(\delta)$.
\end{enumerate}

\subsection{Human Priors as Graph Predicates}
We implement the Human Priors as a set of logic predicates $\mathcal{H} = \{ h_m : \mathcal{V} \times \mathcal{V} \to \{0,1\} \}_{m=1}^M$. For example, the \emph{Boundary Constraint}:
\[
h_{\text{bound}}(i,j) = \mathbb{1}(\text{Brick } i, j \text{ are on the same structural boundary}).
\]
These predicates are injected into the graph by masking the attention matrix in the Transformer:
\[
\mathcal{M}_{ij}^{\text{prior}} = \sum_{m=1}^M w_m \cdot h_m(i,j).
\]
This biases the planner to prioritize logically connected bricks (e.g., finishing a boundary loop) before filling the interior, stabilizing the early phases of training.

\subsection{Optimization Objectives}
\label{OptObj}
EUPHORIA employs an SAC framework adapted for meta-parameter generation.

\textit{1. Critic Update:} The parameters $\omega$ of the soft Q-function are updated by minimizing the Bellman residual:
\begin{align}
    J_Q(\omega) = \mathbb{E}_{\mathcal{B}} \left[ \left( Q_\omega(s,a) - \left(r + \gamma \mathbb{E}_{a' \sim \pi_{\theta}}\left[Q_{\bar{\omega}}(s',a') - \alpha_{\text{ent}} \log \pi_{\theta}(a'|s')\right] \right) \right)^2 \right].
\end{align}

\textit{2. Hypernetwork Update:} Unlike standard SAC where policy parameters $\theta$ are optimized directly, in EUPHORIA, $\theta$ is generated by $\mathcal{H}_\phi(z)$. Thus, we optimize the hypernetwork parameters $\phi$ by backpropagating the policy loss through the generation process:
\begin{align}
    J_\pi(\phi) = \mathbb{E}_{s \sim \mathcal{B}, \epsilon \sim \mathcal{N}} \left[ \alpha_{\text{ent}} \log \pi_{\mathcal{H}_\phi(z)}(f_\phi(\epsilon, s)|s) - Q_\omega(s, f_\phi(\epsilon, s)) \right],
\end{align}
where $f_\phi(\epsilon, s)$ denotes the reparameterized action sample using the generated policy weights $\theta = \mathcal{H}_\phi(z)$. This formulation ensures the Meta-Geometric Encoder learns to generate parameters that intrinsically balance structural stability and energy efficiency across diverse topological tasks.

\subsection{Complexity Analysis and Efficiency}
\label{sec:complexity_analysis}

We analyze the computational complexity of EUPHORIA relative to the baseline categories. Let $N$ be the number of bricks, $E$ the number of edges, $T$ the temporal horizon, and $b$ the branching factor.

\begin{itemize}
    \item \textbf{Search-Based Methods (DFS/BFS/A*):} 
    Traditional search algorithms exhibit exponential time complexity $O(b^d)$ in the worst case, where $d$ is the search depth (sequence length $N$). While they guarantee optimality given an admissible heuristic, they become intractable for large-scale assemblies (e.g., domes with $N>100$), as the physics checks (DEM) must be run for every visited state.

    \item \textbf{Standard GNN Baselines (GCN, GraphSAGE):} 
    Message-passing GNNs typically scale linearly with the number of edges, $O(|E|)$ or $O(N \cdot k)$ where $k$ is the average degree. While computationally cheaper per step than EUPHORIA, they rely on local aggregation (1-hop or 2-hop). To capture global structural stability (e.g., how a keystone affects the base), they require deep stacking, which leads to over-smoothing and performance degradation, as observed in our GCN+RL baselines.

    \item \textbf{Spatiotemporal Baselines (STGCN, ASTGCN):} 
    These models incorporate temporal convolutions, scaling as $O(T \cdot N^2)$ or $O(T \cdot |E|)$. The additional temporal dimension $T$ significantly increases memory/compute requirements during training, making them slower to converge compared to our unified Hypernetwork approach which generates parameters in one shot.

    \item \textbf{EUPHORIA:} 
    Our inference is dominated by the Physics-Informed Graph Transformer mechanism, scaling quadratically $O(N^2)$. Although $O(N^2)$ is heavier than sparse GNNs ($O(|E|)$), it is strictly necessary for modeling dense physical dependencies (force transmission) across the entire structure.
\end{itemize}

\textbf{Efficiency Trade-off.}
EUPHORIA amortizes the $O(N^2)$ cost effectively:
\begin{enumerate}
    \item \textbf{Inference vs. Simulation:} Once trained, generating a full sequence takes seconds (forward passes), whereas A* with physics checks requires re-simulating thousands of branches, taking hours for complex geometries.
    \item \textbf{Sample Efficiency:} The Physics-Bias Attention reduces the effective search space by focusing the policy on structurally critical nodes. Consequently, the SAC agent converges in $\sim$17 hours, whereas standard RL baselines (DQN+GNN) often require days or fail to converge due to sparse rewards.
\end{enumerate}
This trade-off—higher per-step FLOPs for a drastically reduced search horizon and global stability awareness—is justified by the strict real-time constraints of robotic construction.

\section{Control Strategy}
\label{sec:extended_control}

This section details the navigational logic and the rigorous constraint formulation that underpins the EUPHORIA execution layer.

\subsection{Mobile Base Navigational Strategy}
To effectively assemble large-scale structures like domes, the mobile manipulator must reposition itself dynamically. We implement a navigation strategy based on an \textbf{Inverse Reachability Map (IRM)}. The IRM pre-computes valid base poses $\mathbf{q}_{\text{base}} \in SE(2)$ from which a target brick pose $\mathbf{p}_{\text{brick}} \in SE(3)$ is reachable by the arm.
The optimal base position is solved by finding the intersection of feasible regions for a sequence of upcoming bricks, minimizing base movement:
\[
\mathbf{q}_{\text{base}}^* = \underset{\mathbf{x}}{\arg\min} \sum_{k=t}^{t+H} \mathcal{D}(\mathbf{x}, \text{IRM}(\mathbf{p}_{\text{brick}}^{(k)})),
\]
where $\mathcal{D}$ is a distance metric in the configuration space and $H$ is the look-ahead horizon. EUPHORIA's planner outputs not just the brick sequence, but implicitly groups bricks to minimize these base relocations.

\subsection{Constraint Formulation}
While the \textbf{Residual Stability Correction} layer optimizes for energy, it must strictly adhere to the following kinematic constraints $C$, which are checked continuously at $100$Hz during execution:

\begin{itemize}
    \item \textbf{Safety Constraints:}
    \begin{align}
        \|\mathbf{x}(t) - \mathbf{x}_{\text{target}}(t)\| &\leq C_{\text{pos}}, \quad \forall t \in [t_0, T] \\
        \|\dot{\mathbf{x}}(t)\| &\leq C_{\text{vel}}, \quad \|\ddot{\mathbf{x}}(t)\| \leq C_{\text{acc}}
    \end{align}
    \item \textbf{Collision Avoidance:}
    Let $\mathcal{O} = \mathcal{O}_{\text{env}} \cup \mathcal{V}_{\text{placed}}$ be the set of static obstacles and already placed bricks. We enforce:
    \[
    \min_{\mathbf{o} \in \mathcal{O}} \|\mathbf{x}_{\text{robot}}(t) - \mathbf{o}\| > \epsilon_{\text{safe}}, \quad \forall t.
    \]
    Any trajectory violating this is immediately halted, and a replanning request is triggered.
    \item \textbf{Terminal State Compliance:}
    \[
    \lim_{t \to T} \mathbf{x}(t) = \mathbf{x}_{\text{target}} \land \|\boldsymbol{\theta}(t) - \boldsymbol{\theta}_{\text{target}}\| \le C_{\text{orient}}.
    \]
\end{itemize}
Here, $\mathbf{x}(t), \dot{\mathbf{x}}(t), \ddot{\mathbf{x}}(t)$ denote the position, velocity, and acceleration of the end-effector. $C_{\text{pos}}, C_{\text{vel}}, C_{\text{acc}}$ are hardware-specific tolerances.

\subsection{Motion Generation from Graph Sequence}
The discrete sequencing action $a_{\text{seq}}$ produced by the Graph Transformer corresponds to selecting a brick index $\sigma_t$. Specifically, at step $t$, the policy selects:
\[
\sigma_t = \arg\max_{k \in \mathcal{V}^* \setminus \mathcal{V}_t} \pi_\theta(k \mid s_t),
\]
where $\sigma_t$ is the ID of the next brick.
To convert this into motion, we compute the displacement vector $\vec{d} = \mathbf{p}_{\sigma_t} - \mathbf{p}_{\text{current}}$. A time-discretized trajectory is generated:
\[
\mathbf{p}(u) = \mathbf{p}_{\text{current}} + \frac{u}{T} \vec{d}, \quad u \in [0, T],
\]
where $T = \|\vec{d}\| / v_{\text{max}}$. At each time step $u_k$, Inverse Kinematics (IK) is solved, and the collision constraint $C_{\text{collision}}$ is verified. If valid, the Residual Layer then optimizes the fine-grained placement $\delta_t$ around the final target $\mathbf{p}(T)$. This hierarchical approach—Sequence $\to$ Rough Trajectory $\to$ Residual Refinement—ensures efficient and safe execution.

\section{Training Details}
\label{sec:training_details}

Our system utilizes a high-performance workstation equipped with an Intel® Core™ i7-13700KF CPU (5.40 GHz), an Nvidia® GeForce RTX™ 3070 Ti GPU, and 32 GB RAM. The EUPHORIA model is trained via the Kinematics-Aware Soft Actor-Critic (SAC) framework over 200 epochs, with each episode capped at a horizon  corresponding to the number of bricks in the task. All RL-based methods are trained for  episodes with horizon  using a replay buffer  of transitions . For EUPHORIA, the reward is strictly defined in Eq. 11, where  is the refined action and  is predicted from inverse dynamics. Results are reported as mean$\pm$std over 5 random seeds.

\textbf{Task-Specific Conditioning.}
For each training instance (e.g., a \emph{Curve w/ Opening} CAD model), we first extract the support set  to generate task-specific policy weights  via the \textbf{Meta-Geometric Encoder}. Unlike standard RL that trains a fixed policy, EUPHORIA conditions all reward terms on this specific geometry (\textbf{Fig.~\ref{Fig_5}}):
\begin{itemize}
\item \textbf{Task Reward:} Measures the placement accuracy relative to the specific CAD waypoints (e.g., Curve w/ Opening nodes).
\item \textbf{Stability Gate:} Uses the task-specific contact graph to compute DEM force equilibrium.
\item \textbf{Energy Penalty:} Minimizes kinematic costs for the generated topology.
\end{itemize}

\textbf{Optimization Hyperparameters.}
The system is optimized using the Adam optimizer. We employ a two-phase learning rate schedule: an initial rate of  for the Actor/Critic networks and a lower rate of  for the Graph Hypernetwork to ensure stable meta-learning. The learning rate decays by 50\% every 50 epochs. A discount factor  supports long-term structural planning. The entropy coefficient  in SAC is automatically tuned to balance exploration and exploitation. Crucially, the Differentiable Residual Layer is trained end-to-end with the policy using  unrolled gradient descent steps per inference.

\section{Experimental Setup}
\label{sec:exp_setup}

\textbf{Hardware and Software Infrastructure.}
Our experimental setup is robustly designed for simulator-first verification and real-world execution.
\begin{itemize}
\item \textbf{Robotic System:} We utilize a \textbf{Hiwin RA610-1476-GC} robotic arm mounted on a tracked vehicle for enhanced mobility. The system is controlled via \textbf{ROS Noetic} and \textbf{MoveIt}, using Python for high-level logic.
\item \textbf{Simulation \& Physics:} We integrate \textbf{PyTorch} for Graph Reinforcement Learning, \textbf{Compas} for managing Discrete Element Models (DEMs), \textbf{NetworkX} for graph topology, and \textbf{PyBullet} for high-fidelity mobile manipulation simulation.
\item \textbf{Digital Twin:} Real-time visualization and CAD interfacing are handled by \textbf{Rhinoceros 7}.
\end{itemize}

\textbf{Train/Test Splits and Universal Adaptation.}
To rigorously evaluate the \textbf{Meta-Geometric Encoder}, we adopt a strict train/test split based on topological complexity:
\begin{itemize}
\item \textbf{Training Set (Standard Walls):} The model is meta-trained on common planar patterns, such as \textbf{Flemish Bond} and \textbf{Stretcher Bond} layouts.
\item \textbf{Testing Set (Few-Shot):} We evaluate universality on out-of-distribution geometries not seen during training, including \textbf{Domes}, \textbf{Arches}, \textbf{Curved Walls}, and structures with intricate spherical openings.
\end{itemize}
\textbf{Adaptation Protocol:} For these unseen test tasks, we utilize a support set to trigger the Hypernetwork. Crucially, \textbf{no gradient updates} are performed at test time; the task-specific parameters  are generated via a single forward pass.
Our primary evaluation goal is to assess both the \emph{structural validity} of the generated sequence and the \emph{physical precision} of the robotic execution, striving for millimeter-level accuracy (e.g., mm) to validate the efficacy of the Residual Stability Correction layer in real-world applications.

\section{Sim2Real Robustness Details}
\label{sec:sim2real_details}

The transition from simulation to real-world applications is a critical phase in validating the practical viability of our models. Our Sim2Real strategy involves transferring learned behaviors and decision-making processes from highly controlled simulated environments to the variable conditions of physical brick assembly. By leveraging the robust feature extraction of the \textbf{Meta-Geometric Encoder} and the precision of the \textbf{Residual Stability Correction} layer, EUPHORIA facilitates effective adaptation to the discrepancies between simulated and real-world data. We follow the same Sim2Real protocol as in our prior work, injecting Gaussian noise $\mathcal{N}(\mu,\sigma^2)$ into depth and pose estimates during training to mimic empirical sensor deviations. Additionally, domain-randomization techniques perturb lighting, occlusions, and pose estimations to improve robustness against environmental unpredictability.

To minimize the reality gap, we employ the \textbf{ROS Noetic} and \textbf{MoveIt} calibration frameworks on a Hiwin RA610 mobile manipulator. Calibration involves adjusting kinematic parameters based on physical measurements of ArUco Markers placed in the operational environment, utilizing an Eye-in-Hand configuration to refine camera-to-robot transformations. We further synchronize the virtual models with physical movements by positioning the manipulator in 20 diverse poses to refine pose estimation. By integrating real-world sensory data (point clouds, IMU readings) back into the simulation, we reduced the sim–real discrepancy by over 60\%.

\textbf{Impact of Residual Correction.}
A key factor in this success is the \textbf{Residual Stability Correction} layer. Experimental results confirm that this layer significantly reduces the average placement error $e_{\text{mm}}$ compared to the ablation variant \emph{w/o residual correction}. It consistently ensures millimeter-level accuracy by mitigating failure cases caused by small but accumulating geometric and actuation errors that pure sequencing policies often overlook.

\textbf{Scalability and Versatility.}
\textbf{Fig.~\ref{Fig_4}} illustrates the system's scalability in handling complex architectural designs, including \emph{curve w/ opening, double curve, ranking bond, corner, sphere, and dome configurations}. These experiments demonstrate EUPHORIA's capability to adapt to diverse structural challenges, highlighting the graph planner's role in enabling precise placement strategies for varying complexities. Ultimately, the policy successfully executed these intricate shapes with an average placement error of just \textbf{3 mm} and \textbf{zero structural failures} over 20 trials. As shown in \textbf{Fig.~\ref{FS_Fig_3}}, this approach enhances the model's ability to generalize from limited examples, effectively coping with sensing noise and actuation errors absent in pure simulation.

\section{System Challenges}
\label{sec:challenges}

During the development and deployment of EUPHORIA, we addressed three primary challenges inherent to hybrid planning systems:

\begin{enumerate}
  \item \textbf{The Reality Gap (Sim2Real).} High-fidelity CAD models and perfect sensors in simulation often masked critical real-world issues:
  \begin{itemize}
    \item \emph{Sensor Noise and Drift:} Depth and pose estimates fluctuated beyond the Gaussian ranges used in training.
    \item \emph{Unmodeled Dynamics:} Robot–brick interactions exhibited micro-slip behavior not captured by the standard physics engine.
  \end{itemize}
  \emph{Solution:} We introduced the \textbf{Residual Stability Correction} layer. By treating physical refinement as an optimization problem embedded within the policy, the system learns to "expect" these deviations and generates coarse actions that are robustly correctable.

  \item \textbf{Reward Function Refinement.} Designing a reward that balances structural stability with kinematic efficiency proved difficult:
  \begin{itemize}
    \item \emph{Reward Hacking:} Early agents learned to "jiggle" bricks to minimize instantaneous kinematic energy at the expense of long-term structural integrity.
    \item \emph{Sparse Signals:} Long horizons without intermediate stability feedback led to catastrophic forgetting in the RL agent.
    \item \emph{Hyperparameter Sensitivity:} Balancing the weights for stability, energy, and task progress required extensive tuning.
  \end{itemize}
  \emph{Solution:} We implemented the \textbf{Stability Gate} mechanism. By using the DEM simulator as a hard binary filter ($\mathbb{I}(\text{Stable})$), we decoupled stability from optimization, forcing the agent to operate solely within the manifold of physically valid states before optimizing for energy efficiency.

  \item \textbf{Integrating Human Priors.} Purely data-driven policies initially struggled with search space explosion:
  \begin{itemize}
    \item \emph{Implausible Sequences:} The agent would propose attaching bricks in mid-air or in kinematically singular configurations.
    \item \emph{Inefficient Exploration:} Without guidance, the agent ignored common-sense construction logic (e.g., building from corners or ground-up).
  \end{itemize}
  \emph{Solution:} Instead of soft heuristic rewards, we integrated human priors as \textbf{Hard Filters}. By pruning the action space $\mathcal{A}$ using DEM stability checks and IK feasibility validation \emph{before} the network inference, we significantly stabilized training convergence and ensured physically plausible rollouts.
\end{enumerate}

\end{document}